\documentclass[preprint,12pt]{elsarticle}

\usepackage{amssymb}
 \usepackage{amsthm}
\usepackage{amsmath,amsfonts}
\usepackage{graphicx}
\usepackage{booktabs}
\usepackage{longtable}
\usepackage{float}
\usepackage{subfig}
\usepackage{lscape}
\DeclareMathOperator*{\concat}{Concat}
\journal{Information Sciences}

\begin{document}

\begin{frontmatter}


% \ead{hxx@home.swjtu.edu.cn}

\title{Rethinking Attention Mechanism in Time Series Classification}


\author[swjtu]{Bowen Zhao}
\author[swjtu]{Huanlai Xing\corref{cor1}} \ead{hxx@home.swjtu.edu.cn}
\author[swjtu]{Xinhan Wang}\author[swjtu]{Fuhong Song}\author[swjtu]{Zhiwen Xiao}
\cortext[cor1]{Corresponding author}

\affiliation[swjtu]{organization={School of Computing and Artificial Intelligence, Southwest Jiaotong University},%Department and Organization
            city={Chengdu},
            country={China}}

\begin{abstract}
Attention-based models have been widely used in many areas, such as computer vision and natural language processing. However, relevant applications in time series classification (TSC) have not been explored deeply yet, causing a significant number of TSC algorithms still suffer from general problems of attention mechanism, like quadratic complexity. In this paper, we promote the efficiency and performance of the attention mechanism by proposing our flexible multi-head linear attention (FMLA), which enhances locality awareness by layer-wise interactions with deformable convolutional blocks and online knowledge distillation. What's more, we propose a simple but effective mask mechanism that helps reduce the noise influence in time series and decrease the redundancy of the proposed FMLA by masking some positions of each given series proportionally. To stabilize this mechanism, samples are forwarded through the model with random mask layers several times and their outputs are aggregated to teach the same model with regular mask layers. We conduct extensive experiments on 85 UCR2018 datasets to compare our algorithm with 11 well-known ones and the results show that our algorithm has comparable performance in terms of top-1 accuracy. We also compare our model with three Transformer-based models with respect to the floating-point operations per second and number of parameters and find that our algorithm achieves significantly better efficiency with lower complexity.
\end{abstract}




\end{frontmatter}


\section{Introduction}
Nowadays, a huge amount of time series data is stored and analyzed every second in areas like healthcare monitoring, intelligent manufacturing, and many other applications in the Internet of Things \cite{cookAnomalyDetectionIoT2020}\cite{dimartinoIndustrialInternetThings2019}. All these scenarios desire effective time series data mining. In particular, time series classification (TSC) has drawn increasingly more attention in the past few years. A univariate time series is a series of digits that are regularly collected to record the real-time condition of a process for further analysis, for classifying which, a general purpose is to extract temporal representation from the series.

Similar to images in computer vision and sentences in natural language processing, time series data contains global and local patterns \cite{haoNewAttentionMechanism2020}. However, different from images and sentences, a time series consists of continuous and non-stationary data with shapelets \cite{lucasProximityForestEffective2019} that may not be truncated randomly. Different kinds of time series data generally have various original lengths and diverse forms of shapelets. The spans of these shapelets also change even in the same application area. For instance, one can easily recognize whether an object is a dog or a cat only through partial features, like eyes or claws, wherever they are. Nevertheless, a segment of the ECG series is normally meaningless. That's why many techniques, like jigsaw puzzles \cite{liuSelfsupervisedLearningGenerative2021}, are normally powerful to pretrain models in computer vision, but powerless for time series analysis. 

For sequence modeling, long-short term memory networks (LSTMs) can naturally discover temporal features. There are also many attempts that use convolutional neural networks (CNNs) for local feature extraction. Both LSTMs and CNNs have troublesome disadvantages for time series analysis. The former cannot process series data in parallel and may forget meaningful information previously extracted while the latter is usually limited by its inductive bias. Unlike these networks, Transformers allow data to be processed in parallel and have global receptive field, which are widely used in sequence modeling especially for analyzing time series data \cite{russwurmBreizhCropsSatelliteTime}\cite{zerveasTransformerbasedFrameworkMultivariate2021}\cite{jinEndtoendFrameworkCombining2021}\cite{huangResidualAttentionNet2020}. However, Transformers suffer from a complexity of $O(N^2)$ \cite{tayEfficientTransformersSurvey2020} mainly due to the multi-head attention structure adopted. This widely-recognized problem makes it hard to apply to long sequence mining because of the unacceptable time and memory consumption. Therefore, several variants, Linformer \cite{wangLinformerSelfAttentionLinear2020}, Swin Transformer \cite{liuSwinTransformerHierarchical2021}, Swin Transformer V2 \cite{liuSwinTransformerV22021}, Big Bird \cite{zaheerBigBirdTransformers}, Longformer \cite{beltagyLongformerLongDocumentTransformer}, Poolingformer\cite{zhangPoolingformerLongDocument2021}, have been proposed to overcome the problem above in the past few years and their advantages and disadvantages are summarized in \cite{tayEfficientTransformersSurvey2020}. 

To solve different problems, there are two kinds of attention mechanisms, local attention and global attention. In general implementations of local attention mechanisms \cite{liImprovedMultiscaleVision2021}, window-size selection is an unavoidable issue that may destroy the shapelets in a sequence if not addressed properly. Meanwhile, it has been proven that local attention is not superior to those well-designed CNNs in terms of local feature extraction \cite{hanCONNECTIONLOCALATTENTION2022}. Global attention is able to extract features with all positions in the given sequence considered, which makes it well adapted to various time series. To reduce the complexity, several implementations try to approximate the attention maps linearly, like Linformer\cite{wangLinformerSelfAttentionLinear2020}, and SOFT\cite{luSOFTSoftmaxfreeTransformer2021}. They can neither approximate the attention maps accurately nor avoid the noise influence efficiently. Besides, the feature redundancy between multiple heads may also influence the results of classification and hinder the flexibility of an attention-based classification model. To sum up, it is hard for pure Transformer architecture to solve the TSC problem.

According to \cite{fanMaskAttentionNetworks2021}\cite{liEnhancingLocalityBreaking2020}, the original Transformer is designed for global pattern extraction whereas mining local patterns is one of the essential abilities for time series analysis. In TSC, the shapelets hidden in a given time series are normally of various forms and spans. This is why feature extractors require global and local perception. In this regard, there have been many attempts up to now. Some of them designed loosely-coupled Transformer-CNN structures \cite{dascoliConViTImprovingVision2021}\cite{panIntegrationSelfAttentionConvolution2021}\cite{pengConformerLocalFeatures2021} while others implemented Transformers in a CNN-like manner \cite{liuSwinTransformerHierarchical2021}\cite{liuSwinTransformerV22021}. How to enhance Transformers' local feature extraction ability by CNNs becomes one of the most attractive research directions. 

Besides, TSC also suffers from the notorious noise interference in many fields, such as finance \cite{kimFinancialSeriesPrediction2019}, weather \cite{karevanTransductiveLSTMTimeseries2020}, and audio \cite{fanGatedRecurrentFusion2021}. It is easy for human beings to ignore noise because one does not have to analyze a given time series one digit after another which is the actual way for machines. The fluctuations caused by noise may lead to deteriorated performance for local pattern recognition. How to address the noise interference problem is still challenging.

We focus on a new variant of Transformer adapted to TSC problem with less complexity and good performance. To this end, we propose a flexible multi-head linear attention (FMLA) model which can generate high-quality attention maps and avoid noise to some extent. To further reduce the impact of noise and capture useful shapelets, we present a plug-in mask mechanism for model optimization at the training and testing stages. In the meanwhile, we use online distillation to enhance the locality awareness of the proposed FMLA model. Our contributions are summarized below: 
\begin{itemize}
	\item{Our FMLA integrates deformable mechanism proposed in deformable convolutional networks (DCN) \cite{daiDeformableConvolutionalNetworks} into a collaborative linear attention mechanism (CLA) enlightened by \cite{wangLinformerSelfAttentionLinear2020}, \cite{luSOFTSoftmaxfreeTransformer2021} and \cite{liImprovedMultiscaleVision2021}, ensuring accurate approximation to position-wise low-rank attention maps. In CLA, the collaborative attention mechanism \cite{cordonnierMultiHeadAttentionCollaborate2021} is applied for reducing the redundancy and filter out noise between multiple heads. The proposed model realizes lower complexity than the vanilla Transformer model \cite{vaswaniAttentionAllYou2017} in terms of feature extraction.}
	\item{FMLA adopts the mask mechanism to reduce the noise interference by adding random and regular mask layers right after the end of each CLA block. We apply self distillation to random mask layers, which stabilizes and speeds up the training process. Through position-wise random masks in the training process and frequency-based regular masks in the inference process, FMLA avoids local optimum and is more robust when addressing TSC problems.}
	\item{FMLA uses online distillation to circulate those deformable local features captured by DCN blocks within the entire FMLA model, including all the CLA and DCN blocks. Thus, FMLA strengthens its local feature extraction ability via the distillation loss in the backward propagation process. Since the features forwarded are also generated from DCN blocks, we realize forward-backward bidirectional circulation so that FMLA can extract various shapelets from time series data.}
	\item{We compare the proposed model with 11 state-of-the-art TSC algorithms and our model obtains the best accuracy on 36 out of the 85 UCR2018 datasets. What's more, our model ranks first in terms of average rank. Thanks to the linear complexity structure, FMLA achieves significantly better performance with fewer parameters, compared with a number of variants of the vanilla Transformers.}
\end{itemize}
The rest of the paper is organized as follows. In Section 2, we review relevant studies on TSC and Transformers, in particular, how to enhance Transformers with knowledge distillation. The proposed FMLA is detailed in Section 3. In Section 4, performance evaluation and result analysis are provided. Section 5 concludes the work.

\begin{figure*}[!t]
	\centering
	\includegraphics[width=6in]{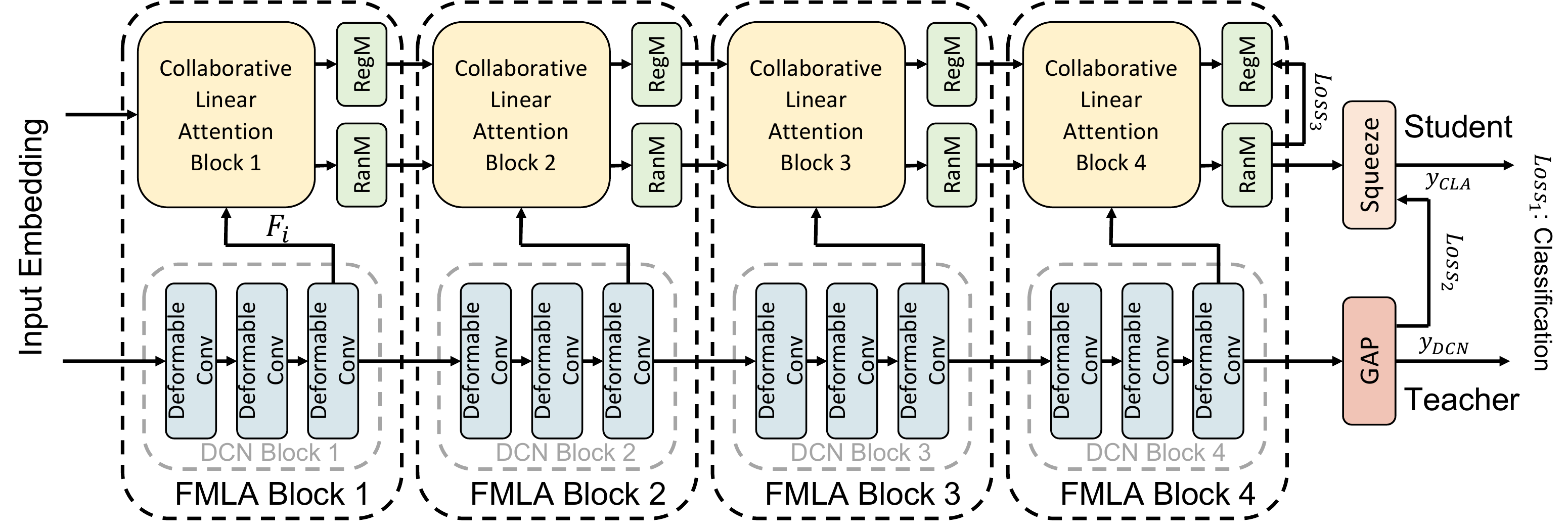}
	\caption{Structure of our algorithm. We use 4 FMLA blocks as an example.}
	\label{pipeline}
\end{figure*}

\section{Related Work}
Firstly, this section reviews traditional algorithms and deep learning ones for TSC. Secondly, a number of well-known variants of Transformer and their trends are given. Thirdly, the existing work on how to enhance Transformer-like structures by knowledge distillation is provided.

\subsection{Traditional and Deep Learning TSC Algorithms}
\textbf{Traditional Algorithms}. Traditional TSC algorithms are usually statistical machine learning based and some important achievements are reviewed below. Lines et al. focused on five temporal modules and introduced HIVE-COTE \cite{linesHIVECOTEHierarchicalVote2016} which integrated several classifiers to hierarchically vote for classification. HIVE-COTE2 \cite{middlehurstHIVECOTENewMeta2021} improved HIVE-COTE by two well-designed classifiers. There have been many other ensemble classifiers, such as Rocket \cite{dempsterROCKETExceptionallyFast2020}, MiniRocket \cite{dempsterMiniRocketVeryFast2021}, and MultiRocket \cite{tanMultiRocketMultiplePooling2022}. Rocket used a number of random convolutional kernels to extract diverse features, all of which then went through a linear classifier to obtain final results. MiniRocket well tuned the hyperparameters and used fixed and small convolutional kernels to speed up the training process. Based on MiniRocket, MultiRocket introduced four additional pooling operators and avoided overfitting by multiple combinations of transformation. Compared with its predecessor, MultiRocket achieved better performance in terms of processing speed and accuracy. Proximity Forest  \cite{lucasProximityForestEffective2019}, like the random forest, made each decision by multiple random proximity trees. This method used parameterized distance measure to compare unclassified samples with each exemplar randomly chosen from each class so that the decision could go through the related branches of a tree. TS-CHIEF \cite{shifazTSCHIEFScalableAccurate2020} was based on the Proximity Forest and organized many advanced classifiers in a tree structure.

\textbf{Deep Learning Algorithms}. With a large number of parameters, deep learning algorithms are capable of capturing detailed and various levels of information. Compared with traditional ones, these algorithms usually achieve better classification results by adaptively learning rich representations from each time series during training. InceptionTime \cite{fawazInceptionTimeFindingAlexNet2020} was a successful application of the Inception networks \cite{szegedyInceptionv4InceptionResNetImpact2016} in time series. However, this model was only good at capturing local patterns due to the Inception block adopted. MACNN \cite{chenMultiscaleAttentionConvolutional2021} used the attention mechanism to improve the classification performance of multi-scale CNNs. Hao et al. presented CA-SFCN \cite{haoNewAttentionMechanism2020}, which adopted variable and temporal attention modules to tackle multi-variate TSC problems. Huang et al. \cite{huangResidualAttentionNet2020} proposed a dual-network-based architecture combining Transformer and ResNet \cite{heIdentityMappingsDeep2016} for TSC, with their outputs concatenated for the final classification. This architecture extracted global and local features separately. Xiao et al. \cite{xiaoRTFNRobustTemporal2021} developed another dual-network architecture, RTFN, placing attention and CNN modules in parallel. Similarly, the features captured by them were concatenated before the classifier. Dual-network architectures are able to extract sufficient and flexible information from a given time series. However, one branch cannot make full use of the hidden states of the other during feature extraction since final classification results are simply generated by concatenating the outputs of the two branches. That motivates us to try layer-wise integration of deformable mechanism and attention mechanism to thoroughly utilize their hidden states, increasing the classification accuracy.

\subsection{Transformer and Its Variants}
\textbf{Low-complexity Transformers.} The vanilla Transformer \cite{vaswaniAttentionAllYou2017} and its application in Computer Vision, ViT  \cite{dosovitskiyImageWorth16x162021}, suffered from quadratic time and space complexity \cite{tayEfficientTransformersSurvey2020}. ViT divided original data into patches which were projected into a low-dimensional space for attention computation. However, it is difficult for a window-based algorithm to always choose suitable window sizes for various lengths of time series. An improper window size may truncate complete shapelets, including those critical ones. This definitely influences the quality of the features extracted since it is hard to analyze truncated features in TSC different from those in computer vision. Therefore, there have been many attempts to deal with the high-complexity and feature truncation problems above. Swin Transformer \cite{liuSwinTransformerHierarchical2021} strengthened its local feature extraction ability by means of a CNN-like architecture. The shift windows technique could, to some extent, relieve the feature truncation problem. However, Swin Transformer also suffered from the problems CNNs encountered. It had to stack more Swin Transformer blocks and required an additional shift layer per block to expand the receptive field \cite{xiaVisionTransformerDeformable2022}. Besides, both ViT and Swin Transformer could not well recognize long-span features. In \cite{liEnhancingLocalityBreaking2020}, Li et al. used convolutional self-attention for local-feature awareness and LogSparse and restart attention mechanisms for memory reduction. Nevertheless, the LogSparse attention might lose useful information while it was hard for the restart attention to choose an appropriate range for a given time series. All the variants above belong to local attention mechanisms. As a global attention mechanism, Linformer \cite{wangLinformerSelfAttentionLinear2020} did not encounter the problems above. It used linear projection to reduce the length of a given input sequence. It was actually a rough approximation to the attention map in the vanilla Transformer model, which definitely led to serious information loss \cite{luSOFTSoftmaxfreeTransformer2021}. In addition, noise data increased the probability that Linformer converged to a local optimum point \cite{fanMaskAttentionNetworks2021}\cite{liEnhancingLocalityBreaking2020}. SOFT \cite{luSOFTSoftmaxfreeTransformer2021} sampled several tokens as queries and achieved a more accurate linear attention through decomposing attention maps. However, it could not reduce the influence in time series. What's worse, one token was usually associated with several others, causing two irrelevant tokens might be identified as related ones in all the attention algorithms mentioned above. Hence, it was possible that the mismatched tokens were then assigned larger weight values wrongly \cite{fanMaskAttentionNetworks2021}.

\textbf{Transformers enhanced by CNNs.} To focus more on the local information, there have been a number of attempts for effectively integrating CNN into Transformer. ConViT \cite{dascoliConViTImprovingVision2021} used gate units in each layer, optimizing the weights of Transformer and CNN blocks in the training process. In this way, the model could adaptively learn the relative importance of each component for different data. ACmix \cite{panIntegrationSelfAttentionConvolution2021} merged the projection processes of Transformer and CNN and then executed the dot-product operations in the former and shift operations in the latter separately. Peng et al. \cite{pengConformerLocalFeatures2021} found that Transformers were not good at capturing local features. The authors constructed a two-branch structure, Transformer and CNN, where the hidden states of one branch were added to the other commutatively for feature supplementation purposes. On the other hand, traditional CNNs have fixed kernel shapes, which cannot extract features with various shapes since local features even in the same class are not always the same under noisy situations. Zhu et al. \cite{zhuDEFORMABLEDETRDEFORMABLE2021} introduced Deformable DERT that generated K coordinates for deformable kernels. For each query vector, its linear transformation was adopted for attention computation. However, K was a fixed hyperparameter, which could not handle problems with various levels of difficulty. Xia et al. proposed a deformable attention module that generated offsets of key and value vectors based on a uniform grid using a lightweight network \cite{zhuDEFORMABLEDETRDEFORMABLE2021}. In canonical attention computation, different queries shared shifted keys and values with each other. The deformable attention module was effective for extracting meaningful features from noisy data. 

\subsection{Knowledge Distillation in Transformers}
Knowledge distillation \cite{hintonDistillingKnowledgeNeural2015} promotes knowledge flows from teacher to student networks, achieving significant parameter compression with acceptable performance degradation. Jiao et al. \cite{jiaoTinyBERTDistillingBERT2020} applied knowledge distillation to the well-known BERT model. The authors distilled on the embedding layer initially and matched the student's layers to the teacher's according to the ratio of the number of layers in the teacher network to that in the student network, where distillation of attention maps and hidden states was implemented between each pair of layers. DistilBERT \cite{sanhDistilBERTDistilledVersion2020} was another attempt on distilled BERT model. The student network had the same architecture as the teacher network but only with half of the layers. DistilBERT achieved similar classification performance as the teacher network and was 60\% faster than the teacher. DeiT \cite{touvronTrainingDataefficientImage} used a different way to distill from the Transformer model. It defined a distillation token that was optimized with all other tokens together just like the CLS token in the BERT model. Experiments showed that it was better for Transformer-like algorithms to use CNN as the teacher. Microsoft raised BERT-PKD \cite{sunPatientKnowledgeDistillation2019} with two distillation strategies, PKD-Last and PKD-Skip. In BERT-PKD, students could either learn from the last K layers or from each K layers.

On the other hand, self-distillation\cite{zhangSelfDistillationEfficientCompact2021}\cite{zhangBeYourOwn2019}, where the teacher and student lie in the same model, has been more and more useful. It can accelerate the training process, compress parameters and even boost the performance. In \cite{liuFastBERTSelfdistillingBERT2020}, Liu et al. proposed FastBERT, where they added the same classifier as the teacher's after each Transformer block and used the output of the last layer to guide the training process of the students. In this way, they realized adaptive inference to reduce the time consumption without performance degradation according to the uncertainty of each level of students. The applications of self-distillation have not been exploited sufficiently. In this paper, we try to use self-distillation to stabilize the mask mechanism in our FMLA.

\begin{figure*}[!t]
	\centering
	\includegraphics[width=6in]{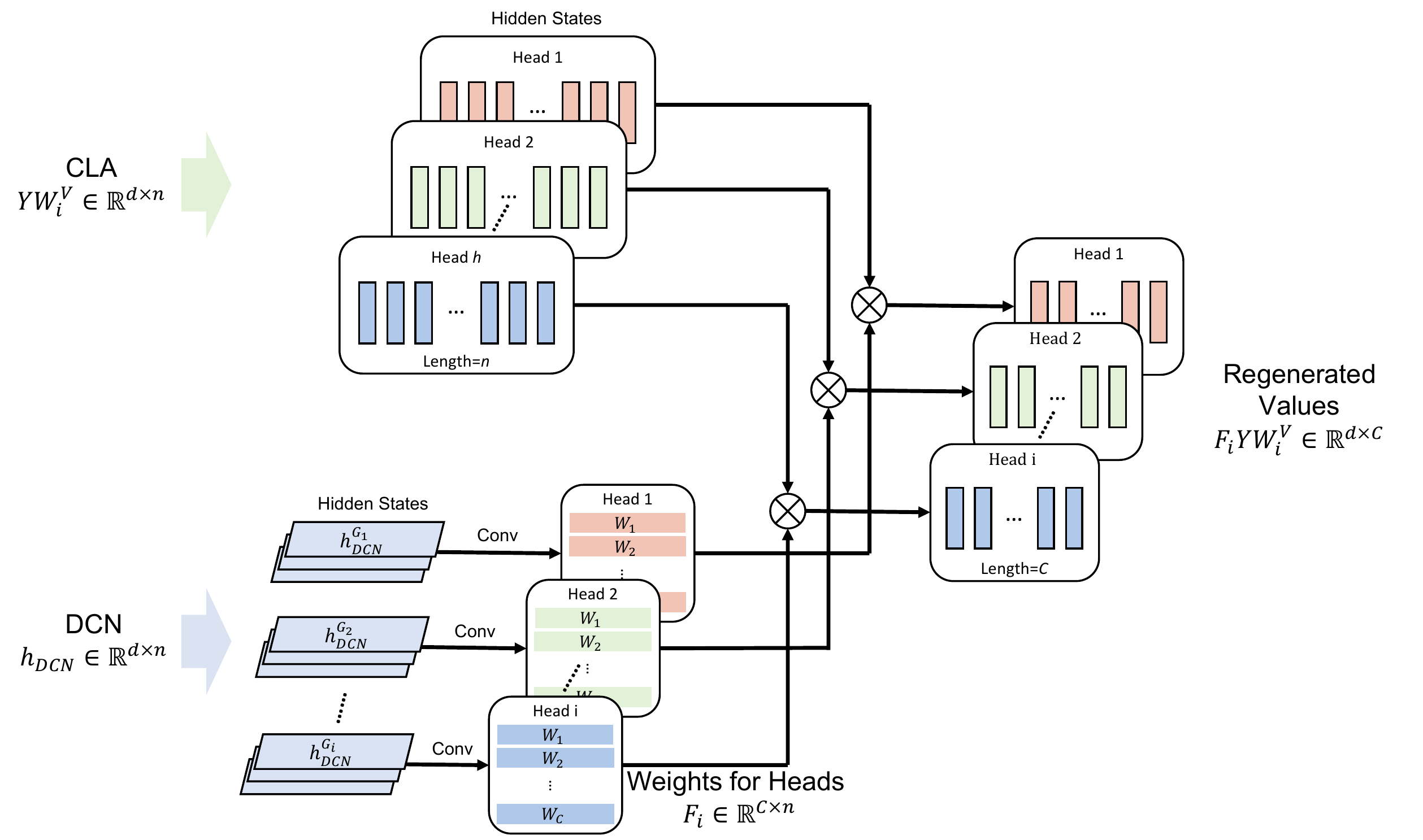}
	\caption{Components of our flexible multi-head linear attention}
	\label{FMLA}
\end{figure*}

\section{Flexible Multi-head Linear Attention (FMLA)}
In this section, we first demonstrate the principle of FMLA through how DCN and CLA blocks interact with each other. Then, we introduce the mask mechanism, including the inspiration and implementation. Finally, the online distillation technique adopted is detailed. The overview of FMLA is shown in Fig. \ref{pipeline}.

\subsection{Flexible Attention Based on Deformable Convolution Networks}\label{3.1}
Existing well-known Transformers, like Swin Transformer \cite{liuSwinTransformerHierarchical2021}, SOFT \cite{luSOFTSoftmaxfreeTransformer2021} and Linformer \cite{wangLinformerSelfAttentionLinear2020}, try to strike a balance between high-complexity computation and effective feature extraction. However, the window-based method in Swin Transformer limits its potential to extract features with various sizes. The sampling technique in SOFT may lose meaningful information with the reduction of complexity. The low-rank assumption and rough approximation in Linformer break the consistency between keys and values. To overcome these problems, we specifically propose a global-local linear attention mechanism, namely FMLA, which realizes accurate attention computation under the guidance of DCN as shown in Fig. \ref{FMLA}. 

FMLA consists of a DCN block and a CLA block in each layer to extract local and global representations with linear complexity. As for a CLA block in Fig. \ref{CLA}, values are regenerated based on the features output by the deformable mechanism and keys are generated based on the new values. We assume that $X$ and $Y$ are the inputs for queries and keys and $X = Y$ in self attention. Eqs. \eqref{att4} and \eqref{att5} redefine the implementation of attention computation and generation of keys in head $i$, $\bar{H_i}$:

\begin{equation}
	\label{att4}
	\bar{H_i} = \text{Attention}(XW_i^\text{Q}, K_i ,F_iYW_i^\text{V}),
\end{equation}
\begin{equation}
	\label{att5}
	K_i=\text{Conv}(F_iV_iW_i^\text{V}),
\end{equation}
where $W_i^\text{Q}$ and $W_i^\text{V}$ are projection matrices. The compressed mapping, $F_i$, is defined in Eq. \eqref{attF}

Since $K_i$ is generated based on $V_i$ through the $\text{Conv()}$ operation with a kernel size of 1 in each head, position-wise relations can be directly constructed and FMLA avoids unnecessary pair-wise inner-product computation. Moreover, there may be several quite different sub-optimal strategies for compression in algorithms like SOFT or Linformer if we train the model on the same dataset several times, which is caused by the instability of sampling or approximation process and is referred to as local optima. Therefore, a proper mapping strategy can reduce model complexity and enable linear attention, i.e. CLA in this paper, to extract more flexible features with less interference. To this end, we project the input sequence in each CLA block based on the output of its counterpart DCN block, $h_{\text{DCN}}^{G_i}$, as shown in 
\begin{equation}
	\label{attF}
	F_i=\text{Conv}_i(h_{\text{DCN}}^{G_i}),
\end{equation}
where $F_i$ is the well-designed projection matrix for head $i$. $\text{Conv}_i()$ is the $i$-th group convolution with a kernel size of 1 that projects feature maps to weights, $F_i$. $G_i$ indexes the input channels of the $i$-th group convolution. $h_{DCN}^{G_i}$ is the hidden states of group $i$ in the counterpart layer of DCN. The number of channels, $C$, in $F_i$ equals the length of compressed values in the CLA block, namely each channel, $W_i$, represents the weights to generate each vector in the regenerated values as shown in Fig. \ref{FMLA}. In this way, each channel group guides one head of the multi-head attention mechanism in each CLA block. As a special branch of CNNs, DCNs are also considered low-level feature extractors. Integrating DCN blocks into CLA blocks enhances their ability to mine local representations and filter noise data.

The linear projections in FMLA clearly lose much information due to the large ratio of compression, which may lead higher layers hard to train. Therefore, we apply the residual pooling connection from MViT \cite{fanMultiscaleVisionTransformers2021}\cite{liImprovedMultiscaleVision2021} to FMLA. Originally, this technique contains pooling operations on the queries, keys and values and the residual connection from queries to the output of the related attention block for reducing the complexity and facilitating the training process. To avoid losing too much information in the linear projection process of the FMLA and minimize the noise influence in the queries, we only add pooled query matrices to the output of each attention block as a low-resolution complement as shown in Eq. \eqref{att6}, where $H_i$ is the output of the $i$-th head and $W_O$ is the compressed mapping of the aggregated information from $N_h$ heads in each FMLA block.

\begin{equation}
	\label{att6}
	\text{MultiHead}(X, Y)= \concat_{i\in{[N_h]}}[\bar{H_i} + \text{Pooling}(XW^\text{Q}_i)]W_O
\end{equation}

In the canonical attention mechanism, each head extracts information individually. The outputs of all heads are concatenated with a squeeze operation for the same-size output, which results in redundant features between heads. To reduce the redundancy, mix vectors are used to represent the unique information of each head before the dot product calculation, as written in Eqs. \eqref{att2} and \eqref{att3}, which is referred to as the collaborative multi-head attention \cite{cordonnierMultiHeadAttentionCollaborate2021}:

\begin{figure}[!t]
	\centering
	\includegraphics[scale=0.5]{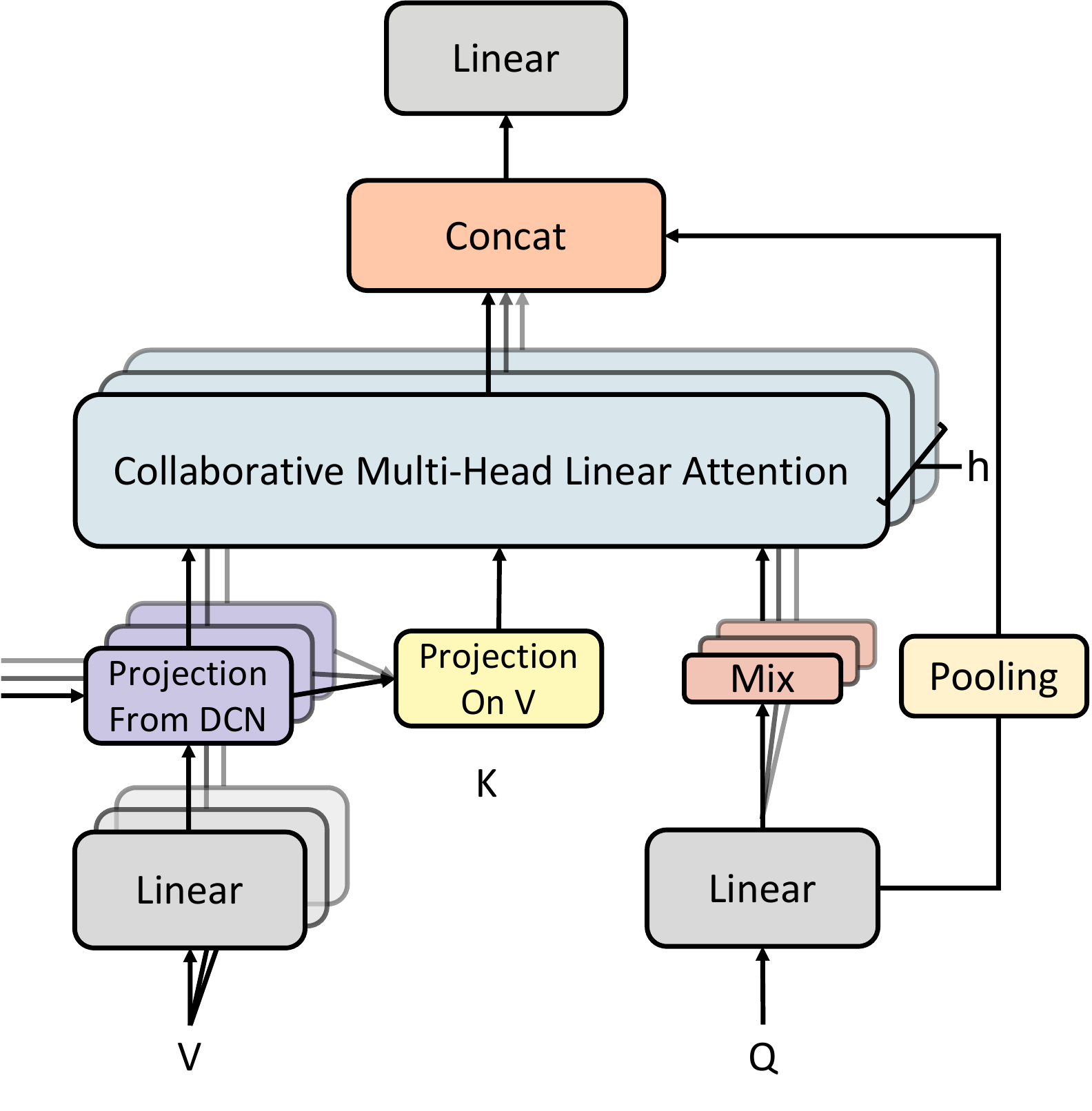}
	\caption{Each CLA block utilizes the compressed mapping generated from the counterpart DCN block. The unique group of key vectors is generated from several groups of values from different heads of the attention mechanism. The mixing vectors originate from the collaborative attention mechanism to help reduce the redundancy in query vectors.}
	\label{CLA}
\end{figure}

\begin{equation}
	\label{att2}
	\bar{H_i}  = \text{Attention}(X\tilde{W^\text{Q}}\text{diag}(m_i), Y\tilde{W}^\text{K}, YW_i^\text{V}),
\end{equation}
\begin{equation}
	\label{att3}
	\text{CollabHead}(X, Y) = \text{Concat}_{i\in{[N_h]}}[\bar{H_i}]W_O,
\end{equation}
where $\tilde{W^Q}$ and $\tilde{W}^K$ are shared across multiple heads. $m_i$ is the mix vector for the unique information extracted in each head and $\text{diag()}$ extends the mix vector to a square matrix.

As known, it is necessary for feature extractors to distinguish noise from shapelets since each head focuses on some of the useful information and unavoidable noise. With the collaboration of multiple heads, a feature extractor can summarize the common features that appear in each head with low noise interference. This is why we implement the collaborative multi-head attention, Eq. \eqref{att2}, in FMLA as Eq. \eqref{att7} , where $m$ is the mix vector for distinguishing the unique knowledge learned by different heads so that projection for each query per head is no longer needed. Since only one key is shared between multiple queries and values in the collaborative attention, we can naturally obtain one key generated by the linear projection from all values as defined in Eq. \eqref{att8}, retaining the position-wise relation between the keys and values. In this way, the attention maps calculated from the dot-product of queries and keys are directly used for weighting the values, $YW_i^V$.

\begin{equation}
	\label{att7}
	H_i = \text{Attention}(X\tilde{W}^Q\text{diag}(m_i),\hat{K},F_iYW_i^V),
\end{equation}
\begin{equation}
	\label{att8}
	\hat{K}=\text{Conv}(\text{Concat}(F_1V_1W_1^V, F_2V_2W_2^V,...,F_iV_iW_h^V)),
\end{equation}
where the shape of $\hat{K}$ is the same as that of $F_iV_iW_i^V$.

\subsection{Mask Mechanism}\label{3.2}
Time series are generally generated with a fixed sampling interval. One can assume that the macroscopic waveforms and critical shapelets are retained as long as we properly lengthen the sampling interval and lower the sampling frequency in most cases. In TSC, there are no such necessary data points that directly affect the classification result since time series specify a continuous process. Recent research \cite{liangNotAllPatches2022} shows that only a few query-key pairs contribute to each related task, which means we may reduce the influence of noise and the redundancy of models by dropping the data in multiple positions of a given time series. In each layer of FMLA, this paper uses position-wise random masks based on Bernoulli distribution in the training process and frequency-based regular masks in the testing process. To be specific, different heads use different random masks and the positions masked are resampled in each iteration. In this way, the FMLA will not overfit specific local features and there is no need to worry about permanent loss of important information. For instance, we randomly mask the data in 50\% of the positions in the sequence, like $X=\{x_1, x_2, 0, 0, x_5, 0, x_7, x_8, 0, 0\}$, during training. In the inference process, the sequence is regularly masked according to the ratio adopted in the training process, i.e. 50\% in the instance above. We mask the data at the second (or first) position of each two consecutive positions during the testing process, like $X=\{x_1, 0, x_3, 0, x_5, 0, x_7, 0, x_9, 0\}$. In this way, low-frequency resampling is achieved. 
The mask technique used in this paper can also be considered as a grouped version of Dropout \cite{srivastavaDropoutSimpleWay}, where we group all weights according to their associated positions and drop those weights in the chosen groups together. That's why the mask mechanism has the ability to resist overfitting (like the original Dropout). Actually, our experimental results show the mask mechanism can work well with Dropout. Details can be found in Subsection \ref{4.2}.

\begin{figure}[t]
	\centering
	\includegraphics[width=2.5in]{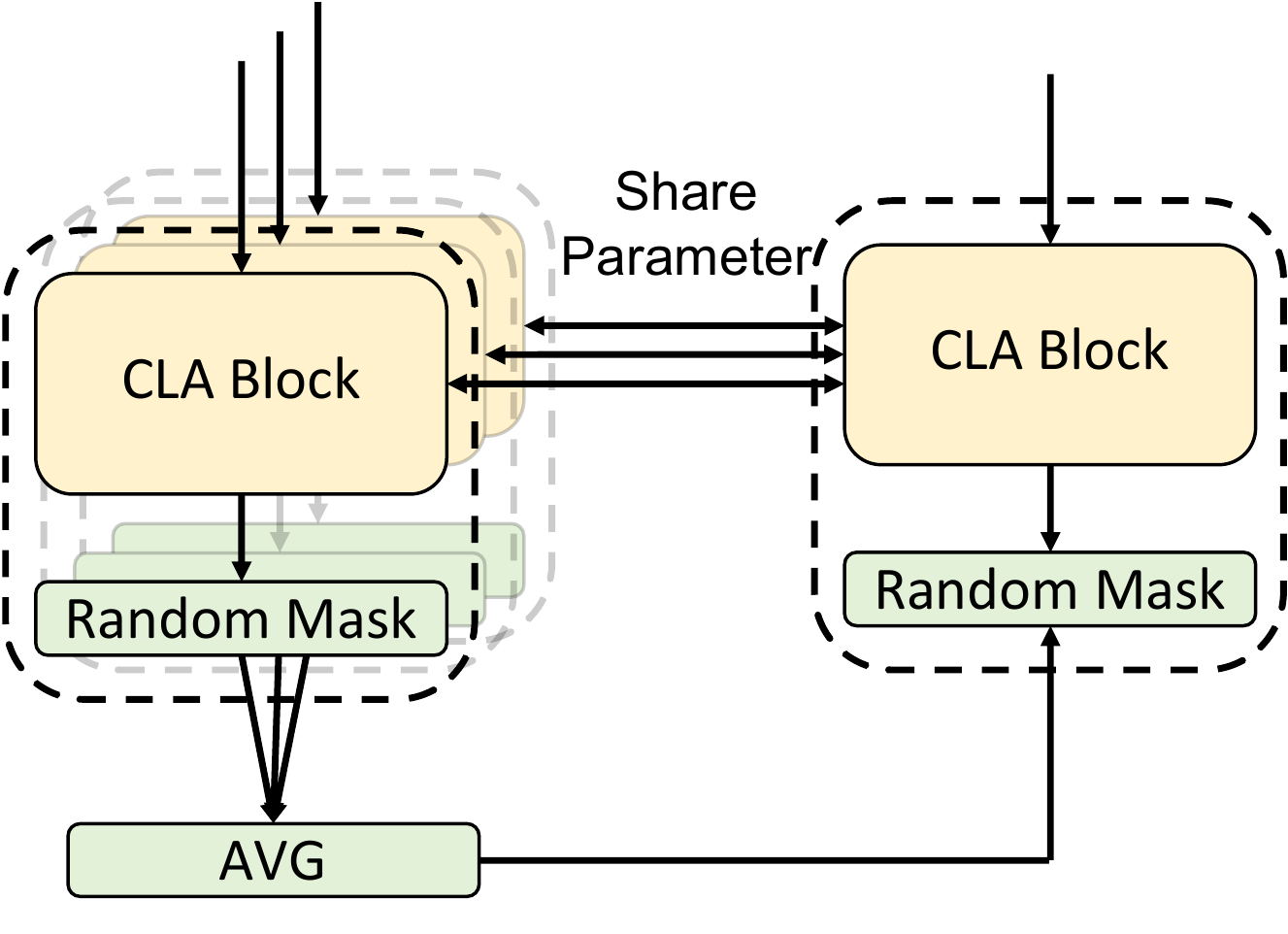}
	\caption{Implementation of self-distillation in the mask mechanism}
	\label{Self-distillation}
\end{figure}

Clearly, our mask mechanism has randomness due to the predefined frequency which may prevent our model from convergence. Self-distillation is invoked to stabilize and speed up the training process as illustrated in Fig. \ref{Self-distillation}. We feed the input time series to the model with random mask layers a predefined number of times. We calculate the average output value and distill the related knowledge to the model above with regular mask layers for the consistency of training and testing processes as shown in Eq. \eqref{loss1}. In this way, we not only stabilize the training process but also make use of the advantages of Bagging in ensemble learning, i.e., variance reduction and random fluctuation stabilization.

\begin{equation}
	\label{loss1}
	\begin{split}
		Loss_1 = \beta \text{D}_\text{KL}(\frac{1}{N}\sum_{t=1}^{N}\text{RandomMask}(\text{FMLA}(x)),\\
		\text{RegularMask}(\text{FMLA}(x))),
	\end{split}
\end{equation}
where $N$ represents a predefined number of times each input time series are fed to the model with random mask layers, $\text{RandomMask()}$.

\subsection{Deformable Attention by Online Distillation}\label{3.3}
According to \cite{touvronTrainingDataefficientImage}, CNN-like algorithms can be excellent teachers for supervising Transformers. Inspired by this, we distill knowledge from the output of DCN and transfer it to the entire FMLA model, i.e., $Loss_2$ in Eq. \eqref{loss2}. Compared with CNN, DCN has better representation ability and can weaken the influence of noise, helping improve the locality awareness of FMLA and concentrate more on deformable features. We utilize the DCN in the FMLA as the teacher. We add the outputs of the last CLA and DCN blocks for classification, which helps retain the independence of DCN. By using DCN blocks to guide the compressed mapping in the forward propagation and adopting the online distillation loss to optimize the backward propagation, we realize a  bidirectional circulation of knowledge mined by DCN, which enhances the vanilla attention mechanism in terms of locality awareness, and noise and complexity reduction.
\begin{equation}
	\label{loss2}
	Loss_2 = \alpha \text{D}_\text{KL}(y_{\text{DCN}}, y_{\text{CLA}}),
\end{equation}
\begin{equation}
	\label{kld}
	\text{D}_\text{KL}(p, q) =\sum_{x\in{X}}p(x)\log\frac{p(x)}{q(x)},
\end{equation}
where $y_{\text{DCN}}$ and $y_{\text{CLA}}$ are the output of two branches. $\text{D}_\text{{KL}}$ means the Kullback-Leibler divergence implemented by Eq. \eqref{kld}. It is used to measure the difference between two distributions, i.e., using distribution $q(x)$ to fit distribution $p(x)$.

\begin{figure}[t]
	\centering
	\includegraphics[width=2.5in]{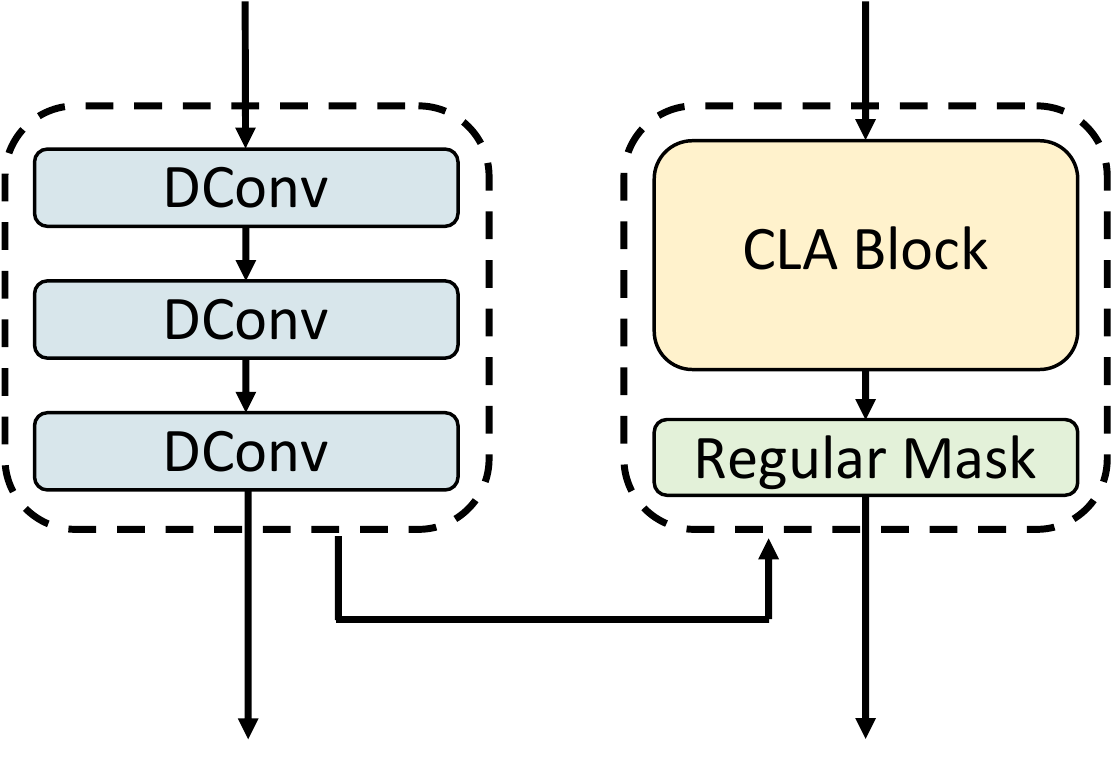}
	\caption{Implementation of online knowledge distillation at the last layers of FMLA and DCN}
	\label{Distillation}
\end{figure}

As the loss decreases during training, the FMLA model is likely to pay more attention to the deformable local features captured by DCN blocks. How much attention is paid can be adjusted by the hyperparameter $\alpha$ in Eq. \eqref{loss2}. In this way, our model achieves a promising perception of global and local patterns. The skeleton of our online distillation method is shown in Fig. \ref{Distillation}.

$Loss_3$ in Eq. \eqref{loss3} is the cross entropy for classification implemented as Eq. \eqref{ce}, and the final loss function consisting of $Loss_1$, $Loss_2$, and $Loss_3$, for optimizing the algorithm shown in Eq. ({\ref{loss}}).
\begin{equation}
	\label{loss3}
	Loss_3 = \text{H}(\hat{y}, y)
\end{equation}
\begin{equation}
	\label{ce}
	\text{H}(p, q) = \sum_{i=1}^{n}p(x_i)log(q(x_i))
\end{equation}

\begin{equation}
	\label{loss}
	Loss = Loss_1 + Loss_2 + Loss_3
\end{equation}

\section{Experiments and Analysis}
We first introduce the experimental setup and ablate the three important improvements of the FMLA, including the DCN-enhanced multi-head attention, mask mechanism, and online distillation. Then, the complexity of our model and its comparison with a number of algorithms is analyzed.
\subsection{Experimental Setup}
We evaluate the performance of the FMLA model by comparing it with 11 state-of-the-art TSC algorithms including \cite{huangResidualAttentionNet2020} against `win'/`tie'/`lose', MeanACC, and AVG\_rank based on top-1 accuracy on 85 out of the 128 UCR2018 datasets \cite{dauUCRTimeSeries2019}. Specifically, the `win' indicates that on how many datasets an algorithm achieves the unique best results compared with all other algorithms. The `tie' means on how many datasets the algorithm gets the same results as the best ones. The `lose' represents on how many datasets the algorithm does not achieve the best results. The number of `best' is the summation of the number of cases of `win' and `tie' for each algorithm. `MeanAcc' is used to compare the general accuracy of different algorithms and `AVG\_Rank' is the result of the Friedman test for ranking all algorithms in the whole dataset averagely.
The 85 datasets are composed of 65 `short' or `medium' and 20 `long' time series datasets. `Existing SOTA', `TS-CHIEF' and `MACNN' only have publically available benchmark results on 65 datasets for performance comparison. There are four FMLA blocks in our experiments, where each block normally has 4 heads. We compress the length of each input sequence to 16 by the projection matrices generated by each DCN block for most cases. We mask the data in 50\% of the positions of the output after each CLA block. As for activation functions, we use the ReLU function in DCN blocks and GELU \cite{hendrycksGaussianErrorLinear2020} in each CLA block. We use 128 kernels in the first two DCN blocks, 64 in the last two as \cite{huangResidualAttentionNet2020} suggested and set the kernel size to 3 for feature extraction and position generation layers. Note that the parameter settings above are used in the ablation study and overall performance comparison. There may be small adjustments of hyperparameter setting in a few special datasets, e.g., for complex and long input sequences, we utilize more heads in the attention mechanism and higher ratios in the mask mechanism.

%\begin{landscape}
	{
		\small\tiny
\begin{longtable}{@{}cccccc@{}}
	\caption{The top-1 accuracy results of the 5 models on 85 UCR2018 datasets.} \label{ablation} \\
	\toprule
	Dataset               & ResNet-Transformer & CLAwDCN  & CLAwDCN-M & CLAwDCN-M-SD & FMLA     \\ \midrule
	Adiac                 & 0.849105           & 0.810742 & 0.808184  & 0.836317     & 0.85422  \\
	ArrowHead             & 0.891429           & 0.771429 & 0.834286  & 0.834286     & 0.857143 \\
	Beef                  & 0.866667           & 0.633333 & 0.8       & 0.7          & 0.833333 \\
	BeetleFly             & 1                  & 0.8      & 0.85      & 1            & 1        \\
	BirdChicken           & 0.95               & 0.85     & 0.95      & 1            & 1        \\
	Car                   & 0.866667           & 0.883333 & 0.9       & 0.9          & 0.916667 \\
	CBF                   & 1                  & 0.987778 & 0.997778  & 0.998889     & 1        \\
	ChlorineCon.          & 0.409375           & 0.782813 & 0.8125    & 0.829948     & 0.815365 \\
	CinCECGTorso          & 0.89058            & 0.902899 & 0.782609  & 0.805797     & 0.832609 \\
	Coffee                & 1                  & 1        & 1         & 1            & 1        \\
	CricketX              & 0.810256           & 0.751282 & 0.771795  & 0.738462     & 0.787179 \\
	CricketY              & 0.825641           & 0.769231 & 0.771795  & 0.776923     & 0.794872 \\
	CricketZ              & 0.128205           & 0.764103 & 0.761538  & 0.761538     & 0.8      \\
	DiatomSizeRe.         & 0.379085           & 0.630719 & 0.934641  & 0.937908     & 0.996732 \\
	DistalPhalanxO.A.G    & 0.467626           & 0.769784 & 0.798561  & 0.798561     & 0.805755 \\
	DistalPhalanxO.C.     & 0.822464           & 0.797101 & 0.793478  & 0.822464     & 0.826087 \\
	Earthquakes           & 0.76259            & 0.798561 & 0.776978  & 0.805755     & 0.798561 \\
	ECG200                & 0.94               & 0.92     & 0.93      & 0.93         & 0.96     \\
	ECG5000               & 0.944222           & 0.942    & 0.939778  & 0.945333     & 0.944222 \\
	ECGFiveDays           & 1                  & 1        & 0.930314  & 0.974448     & 0.977933 \\
	FaceFour              & 0.977273           & 0.806818 & 0.943182  & 0.795455     & 0.977273 \\
	FacesUCR              & 0.926829           & 0.949268 & 0.952683  & 0.953659     & 0.954146 \\
	FordA                 & 0.517424           & 0.940152 & 0.935606  & 0.935606     & 0.939394 \\
	FordB                 & 0.838272           & 0.823457 & 0.828395  & 0.822222     & 0.82716  \\
	GunPoint              & 1                  & 1        & 1         & 1            & 1        \\
	Ham                   & 0.619048           & 0.809524 & 0.938095  & 0.809524     & 0.838095 \\
	HandOutlines          & 0.835135           & 0.948649 & 0.943243  & 0.951351     & 0.959459 \\
	Haptics               & 0.600649           & 0.519481 & 0.535714  & 0.564935     & 0.564935 \\
	Herring               & 0.65625            & 0.703125 & 0.703125  & 0.75         & 0.71875  \\
	InsectWingbeatS.      & 0.535859           & 0.636869 & 0.461616  & 0.532323     & 0.835341 \\
	ItalyPowerDemand      & 0.962099           & 0.96793  & 0.969874  & 0.969874     & 0.971817 \\
	Lightning2            & 0.754098           & 0.885246 & 0.836066  & 0.836066     & 0.901639 \\
	Lightning7            & 0.383562           & 0.821918 & 0.780822  & 0.849315     & 0.849315 \\
	Mallat                & 0.934328           & 0.882729 & 0.945416  & 0.957783     & 0.975693 \\
	Meat                  & 1                  & 0.916667 & 0.983333  & 0.983333     & 1        \\
	MedicalImages         & 0.759211           & 0.767105 & 0.761842  & 0.756579     & 0.756579 \\
	MiddlePhalanxO.A.G.   & 0.623377           & 0.662338 & 0.655844  & 0.675325     & 0.668831 \\
	MiddlePhalanxO.C.     & 0.848797           & 0.862543 & 0.848797  & 0.872852     & 0.869416 \\
	MiddlePhalanxTW       & 0.551948           & 0.580909 & 0.590909  & 0.61039      & 0.623377 \\
	MoteStrain            & 0.9377             & 0.861821 & 0.901757  & 0.881789     & 0.930511 \\
	OliveOil              & 0.933333           & 0.933333 & 0.933333  & 0.933333     & 0.966667 \\
	Plane                 & 0.371492           & 1        & 1         & 1            & 1        \\
	ProximalPhalanxO.A.G. & 0.882927           & 0.897561 & 0.887805  & 0.892683     & 0.897561 \\
	ProximalPhalanxO.C.   & 0.683849           & 0.924399 & 0.931271  & 0.927835     & 0.938144 \\
	ProximalPhalanxTW     & 0.819512           & 0.84878  & 0.84878   & 0.853659     & 0.853659 \\
	ShapeletSim           & 0.888889           & 0.922222 & 0.955556  & 1            & 1        \\
	ShapesAll             & 0.921667           & 0.928333 & 0.93      & 0.921667     & 0.933333 \\
	SonyAIBORobotSur.1    & 0.708819           & 0.960067 & 0.956739  & 0.976705     & 0.985025 \\
	SonyAIBORobotSur.2    & 0.98426            & 0.940189 & 0.951731  & 0.965373     & 0.965373 \\
	Strawberry            & 0.986486           & 0.986486 & 0.983784  & 0.986486     & 0.989189 \\
	SwedishLeaf           & 0.9696             & 0.9664   & 0.9584    & 0.9632       & 0.968    \\
	Symbols               & 0.976884           & 0.849246 & 0.935678  & 0.955779     & 0.982915 \\
	SyntheticControl      & 1                  & 1        & 1         & 1            & 1        \\
	ToeSegmentation1      & 0.97807            & 0.850877 & 0.872807  & 0.921053     & 0.982456 \\
	ToeSegmentation2      & 0.953846           & 0.892308 & 0.907692  & 0.938462     & 0.938462 \\
	Trace                 & 1                  & 1        & 1         & 1            & 1        \\
	TwoLeadECG            & 1                  & 0.979807 & 1         & 1            & 1        \\
	TwoPatterns           & 1                  & 1        & 1         & 1            & 1        \\
	UWaveGestureL.All     & 0.939978           & 0.892797 & 0.919598  & 0.904802     & 0.893076 \\
	UWaveGestureL.X       & 0.810999           & 0.789782 & 0.800391  & 0.805974     & 0.805974 \\
	UWaveGestureL.Y       & 0.671413           & 0.687046 & 0.68928   & 0.698492     & 0.698492 \\
	UWaveGestureL.Z       & 0.760469           & 0.758515 & 0.735064  & 0.719314     & 0.738693 \\
	Wafer                 & 0.99854            & 0.99708  & 0.993835  & 0.993511     & 0.99562  \\
	Wine                  & 0.87037            & 0.851852 & 0.907407  & 0.814815     & 0.962963 \\
	WordSynonyms          & 0.636364           & 0.564263 & 0.559561  & 0.540752     & 0.592476 \\
	ACSF1                 & 0.93               & 0.91     & 0.91      & 0.9          & 0.93     \\
	BME                   & 1                  & 0.973333 & 1         & 0.986667     & 1        \\
	Chinatown             & 0.985507           & 0.976676 & 0.982507  & 0.988338     & 0.988338 \\
	Crop                  & 0.746012           & 0.72869  & 0.725238  & 0.705952     & 0.725119 \\
	DodgerLoopDay         & 0.4625             & 0.575    & 0.575     & 0.575        & 0.6125   \\
	DodgerLoopGame        & 0.550725           & 0.818841 & 0.92029   & 0.913043     & 0.905797 \\
	DodgerLoopWeekend     & 0.949275           & 0.971014 & 0.971014  & 0.978261     & 0.985507 \\
	GunPointAgeSpan       & 1                  & 0.993671 & 0.996835  & 1            & 1        \\
	GunPointMaleV.F.      & 0.996835           & 1        & 1         & 1            & 1        \\
	GunPointOldV.Y.       & 1                  & 1        & 1         & 0.993651     & 1        \\
	InsectEPGRegularT.    & 1                  & 0.911647 & 0.947791  & 1            & 1        \\
	InsectEPGSmallT.      & 0.971888           & 0.714859 & 0.823293  & 0.763052     & 0.919679 \\
	MelbournePedestrian   & 0.904898           & 0.899549 & 0.893399  & 0.895449     & 0.889299 \\
	PowerCons             & 0.927778           & 0.938889 & 0.961111  & 0.977778     & 0.977778 \\
	Rock                  & 0.82               & 0.88     & 0.8       & 0.78         & 0.78     \\
	SemgHandG.Ch2         & 0.848333           & 0.836667 & 0.818333  & 0.845        & 0.87     \\
	SemgHandM.Ch2         & 0.391111           & 0.511111 & 0.431111  & 0.511111     & 0.533333 \\
	SemgHandS.Ch2         & 0.666667           & 0.831111 & 0.713333  & 0.831111     & 0.8      \\
	SmoothSubspace        & 0.993333           & 0.986667 & 0.986667  & 1            & 1        \\
	UMD                   & 1                  & 1        & 1         & 1            & 1        \\
	\midrule
	MeanAcc			& 0.82304	& 0.856361471	& 0.867943388	& 0.873686094	& 0.893739259 \\
	\bottomrule
\end{longtable}
}

\subsection{Ablation Study}\label{4.2}
To validate the effectiveness of the techniques proposed in this paper, we conduct an incremental ablation study with 85 UCR2018 datasets. The incremental models are listed below.

- ResNet-Transformer: a well-known dual-network model that consists of a ResNet branch and a vanilla Transformer branch, i.e., vanilla Transformer-based model \cite{huangResidualAttentionNet2020}.

- CLAwDCN: the collaborative linear attention with DCN integrated, in Subsection \ref{3.1}.

- CLAwDCN-M: CLAwDCN with mask mechanism in Subsection \ref{3.2}. 

- CLAwDCN-M-SD: CLAwDCN-M with self-distillation in Subsection \ref{3.2}.

- FMLA: CLAwDCN-M-SD with online distillation in Subsection \ref{3.3}, i.e., the proposed model in this paper.

\begin{figure}
	\centering
	\includegraphics[width=2.5in]{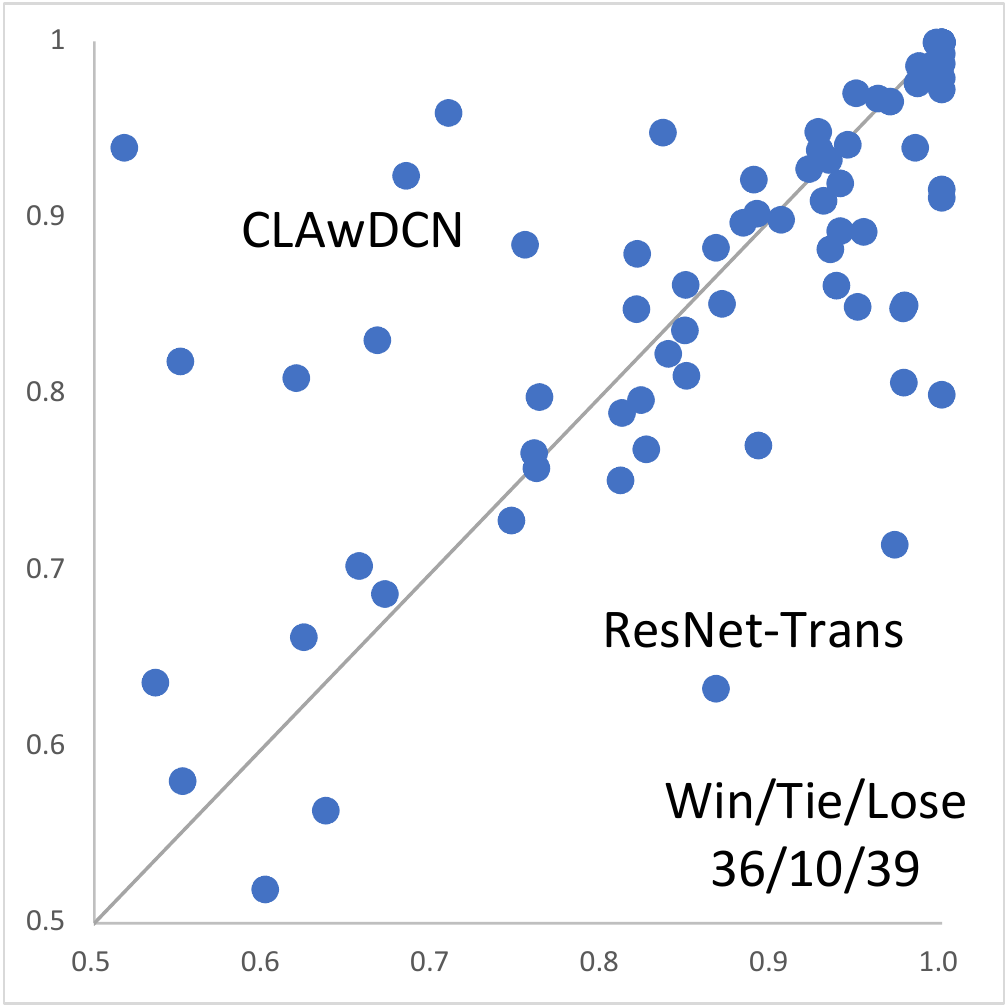}
	\caption{Accuracy plot showing the performance difference between ResNet-Transformer and CLAwDCN.}
	\label{ablation1}
\end{figure}

In terms of accuracy, the performance comparison between CLAwDCN and ResNet-Transformer \cite{huangResidualAttentionNet2020} is shown in Fig. \ref{ablation1}. The two models have the same structure. CLAwDCN and ResNet-Transformer are close runners with 85 datasets considered. CLAwDCN promotes the mean accuracy to about 0.856163 as shown in Table \ref{ablation}. CLAwDCN wins in 36 cases and loses in 39 cases, reflecting that CLAwDCN does not lead to performance deterioration, compared with the vanilla Transformer-based model. ResNet-Transformer extracts local and global features separately, which is actually an advanced voting algorithm based on two kinds of features. Different from CNN, DCN can emphasize local features with various shapes and the compressed matrices for each head are generated by different channel groups in the associated DCN block. Thus, CLAwDCN can calculate the similarity between useful data points based on the inputs filtered by DCN blocks, which helps avoid the influence of fluctuation and capture more accurate information rather than the noise in time series.

\begin{figure}
	\centering
	\includegraphics[width=2.5in]{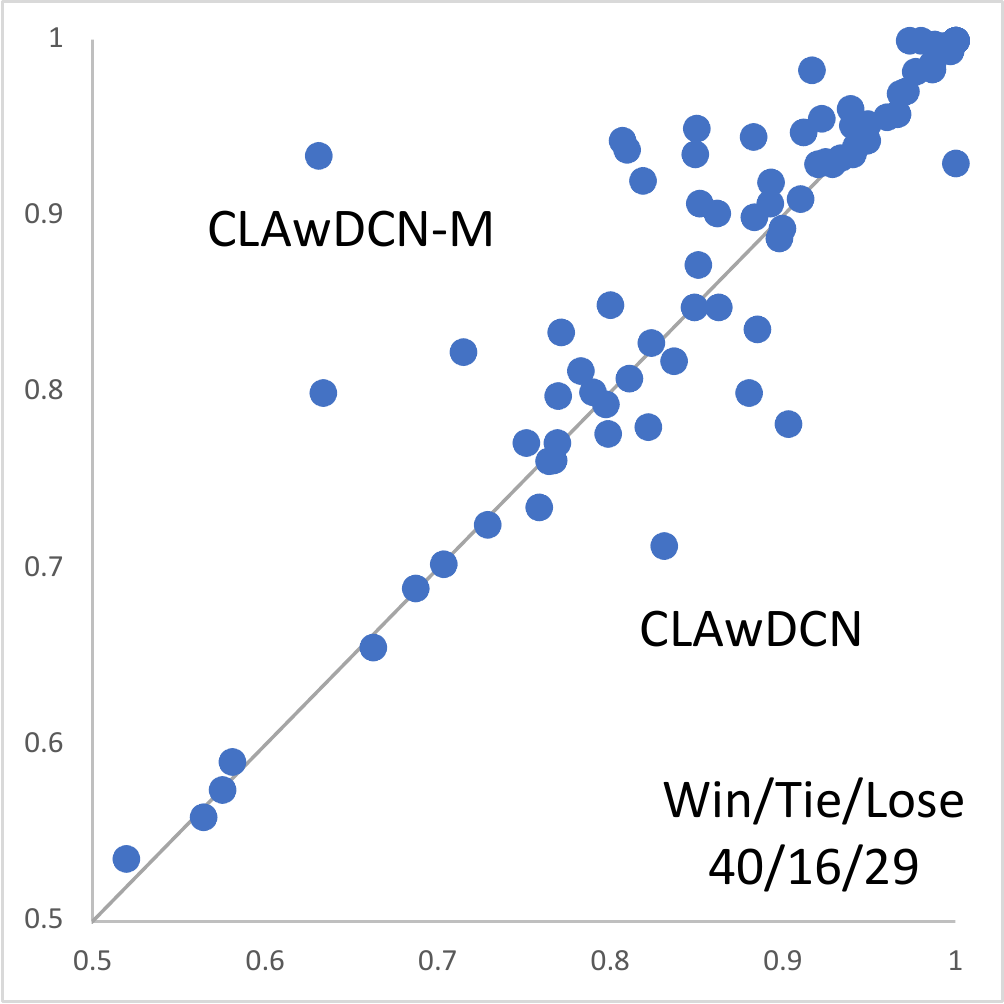}
	\caption{Accuracy plot showing the performance difference between CLAwDCN and CLAwDCN-M.}
	\label{ablation2}
\end{figure}

Then, we evaluate the effectiveness of the mask mechanism without self-distillation by comparing CLAwDCN-M and CLAwDCN with respect to their performance on accuracy. In theory, different mask frequency values can be used in the shallow and deep layers of CLAwDCN-M. For simplicity purposes, we only adopt the same frequency, i.e., 50\%, in all layers in the experiment. The mean accuracy was promoted by CLAwDCN-M to about 0.873686 as shown in Table \ref{ablation}. The performance comparison between CLAwDCN-M and CLAwDCN is shown in Fig. \ref{ablation2}, where the former wins 40 cases and loses 29 cases. If we only use the Dropout \cite{srivastavaDropoutSimpleWay} technique which randomly drops the weights of neural networks, the output of each position the weights associated with still exists and the relevant influence also remains. In addition, we speculate that in the attention mechanism the classification-related data generally obtains higher weights after the dot product operation, which means that even if some specific positions are dropped, they may also remain in other positions in the form of a weighted summation.

\begin{figure}
	\centering
	\includegraphics[width=2.5in]{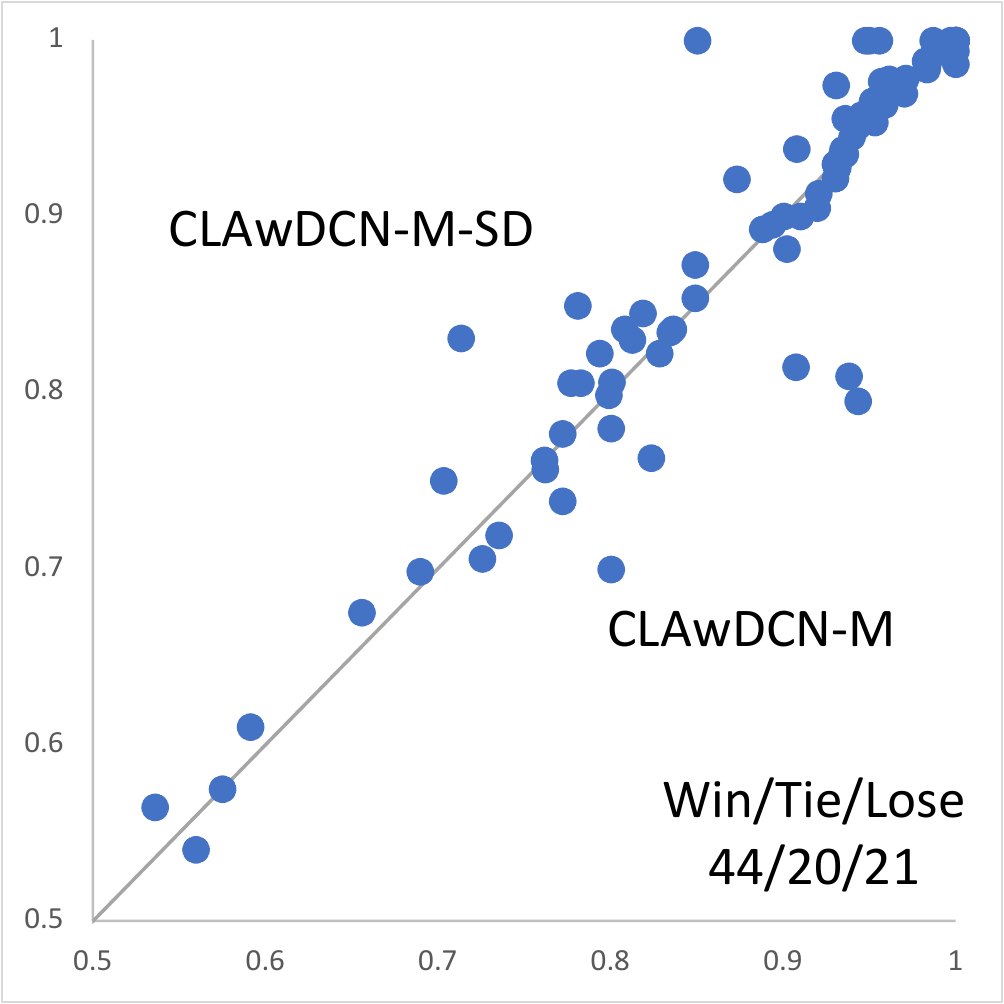}
	\caption{Accuracy plot showing the performance difference between CLAwDCN-M and CLAwDCN-M-SD.}
	\label{ablation3}
\end{figure}

As mentioned in Subsection \ref{3.2}, self-distillation is applied to the mask mechanism to stabilize the training process. We evaluate the effectiveness of self-distillation by comparing CLAwDCN-M-SD and CLAwDCN-M regarding accuracy. We feed the input time series to CLAwDCN-M with random mask layers 3 times. The averaged output is used to teach the same model with regular mask layers, i.e., CLAwDCN-M-SD. The mean accuracy of CLAwDCN-M-SD is boosted to 0.873686 by self-distillation as shown in Table \ref{ablation}. The accuracy plots of CLAwDCN-M-SD vs. CLAwDCN-M are shown in Fig. \ref{ablation3}. CLAwDCN-M-SD wins CLAwDCN-M in 44 cases and loses in 21 cases. The mask mechanism can reduce the influence of noise, and the data in the masked positions may not break the whole waveform of the time series ideally. However, mask operations may appear at different positions in different iterations. This may enhance the model's robustness but makes it hard to train. The self-distillation technique used here is for accelerating the training process under randomness. 

\begin{figure}
	\centering
	\includegraphics[width=2.5in]{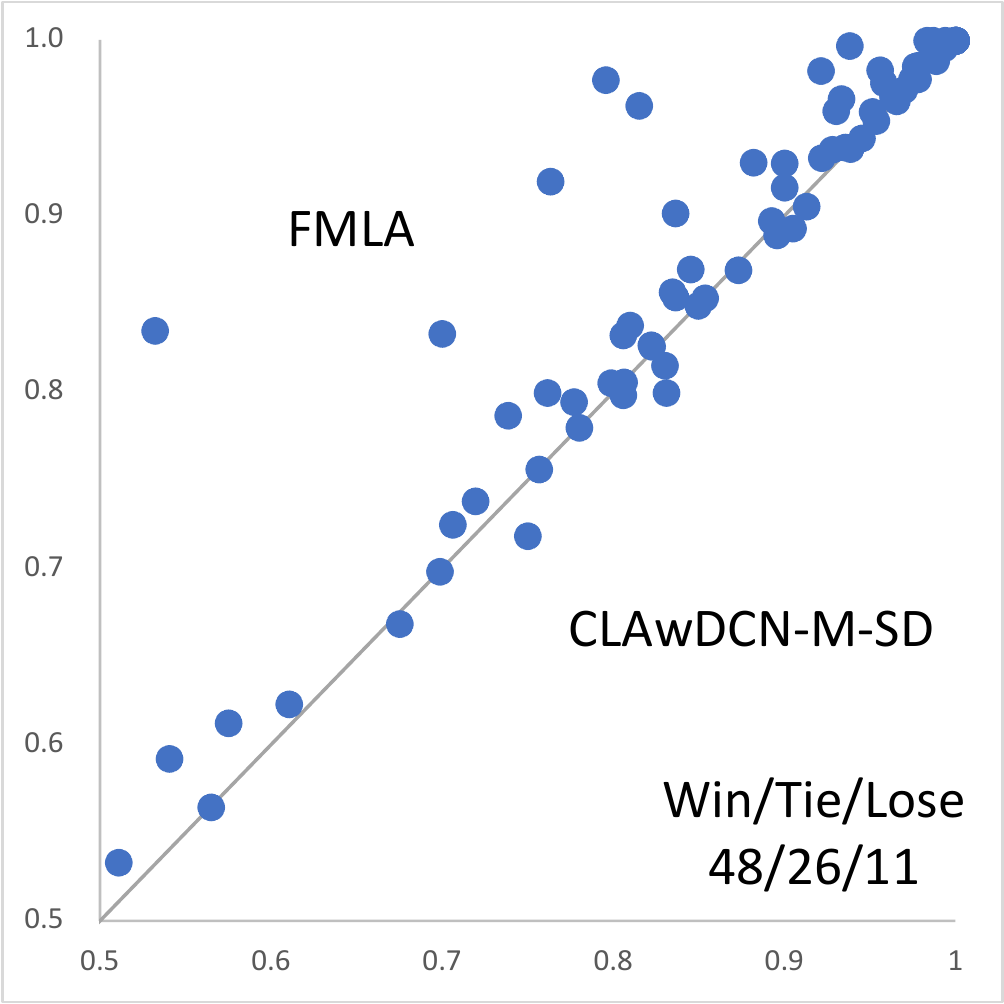}
	\caption{Accuracy plot showing the performance difference between CLAwDCN-M-SD and FMLA.}
	\label{ablation4}
\end{figure}

Finally, we evaluate the effectiveness of the online distillation by comparing CLAwDCN-M-SD and FMLA. Obviously, FMLA overweighs CLAwDCN-M-SD in terms of `win'/`tie'/`lose' as shown in Fig. \ref{ablation4}. The final accuracy of FMLA is 0.893739 and it is the winner of 48 cases and the loser of 11 cases. Online distillation helps realize the forward-backward bi-directional circulation in our FMLA. Therefore, the locality awareness of attention in the FMLA is further strengthened. This also proves the proposition in \cite{touvronTrainingDataefficientImage} that CNN-like architectures can be good teachers for Transformer-like algorithms in knowledge distillation.

{
	\small\tiny
	\setlength{\tabcolsep}{2pt}
\begin{landscape}
	\begin{longtable}[]{@{}cccccccccccccc@{}}
	\caption{The top-1 accuracy results of the 12 models on 85 UCR2018 datasets.}\label{overall} \\
				\toprule
				Dataset               & Existing SOTA & Best:LSTM-FCN & Vanilla:Trans & ResNet-Trans1 & ResNet-Trans2 & ResNet-Trans3 & TS-CHIEF & ResNet-50 SC & Inception   & ROCKET     & MACNN & FMLA      \\ \midrule
				Adiac                 & 0.857         & 0.869565      & 0.84399             & 0.849105      & 0.849105      & 0.849105      & 0.798    & 0.844        & 0.841432225 & 0.78337596 & 0.824 & 0.85422  \\
				ArrowHead             & 0.88          & 0.925714      & 0.891429            & 0.891429      & 0.891429      & 0.897143      & 0.8327   & 0.8857       & 0.845714286 & 0.81428571 & 0.863 & 0.857143 \\
				Beef                  & 0.9           & 0.9           & 0.866667            & 0.866667      & 0.866667      & 0.866667      & 0.7061   & 0.7333       & 0.7         & 0.83333333 & 0.933 & 0.833333 \\
				BeetleFly             & 0.95          & 1             & 1                   & 0.95          & 1             & 0.7           & 0.9136   & 0.9          & 0.8         & 0.9        & 1     & 1        \\
				BirdChicken           & 0.95          & 0.95          & 1                   & 0.9           & 0.95          & 1             & 0.9091   & 0.9          & 0.95        & 0.9        & 1     & 1        \\
				Car                   & 0.933         & 0.966667      & 0.95                & 0.883333      & 0.866667      & 0.3           & 0.8545   & 0.8833       & 0.883333333 & 0.84666667 & 0.917 & 0.916667 \\
				CBF                   & 1             & 0.996667      & 1                   & 0.997778      & 1             & 1             & 0.9979   & 0.9044       & 0.998888889 & 1          & 1     & 1        \\
				ChlorineCon.          & 0.872         & 0.816146      & 0.849479            & 0.863281      & 0.409375      & 0.861719      & 0.7167   & 0.7844       & 0.8765625   & 0.81453125 & 0.888 & 0.815365 \\
				CinCECGTorso          & 0.9949        & 0.904348      & 0.871739            & 0.656522      & 0.89058       & 0.31087       & 0.9832   & 0.8913       & 0.853623188 & 0.83615942 & 0.886 & 0.832609 \\
				Coffee                & 1             & 1             & 1                   & 1             & 1             & 1             & 1        & 1            & 1           & 1          & 1     & 1        \\
				CricketX              & 0.821         & 0.792308      & 0.838462            & 0.8           & 0.810256      & 0.8           & 0.8138   & 0.7359       & 0.853846154 & 0.81948718 & 0.862 & 0.787179 \\
				CricketY              & 0.8256        & 0.802564      & 0.838462            & 0.820513      & 0.825641      & 0.807692      & 0.8019   & 0.7359       & 0.851282051 & 0.85230769 & 0.869 & 0.794872 \\
				CricketZ              & 0.8154        & 0.807692      & 0.820513            & 0.805128      & 0.128205      & 0.1           & 0.834    & 0.7282       & 0.861538462 & 0.85589744 & 0.121 & 0.8      \\
				DiatomSizeRe.         & 0.967         & 0.970588      & 0.993464            & 0.996732      & 0.379085      & 0.996732      & 0.973    & 0.9379       & 0.934640523 & 0.96993464 & 0.977 & 0.996732 \\
				DistalPhalanxO.A.G    & 0.835         & 0.791367      & 0.81295             & 0.776978      & 0.467626      & 0.776978      & 0.7462   & 0.7842       & 0.73381295  & 0.75899281 & 0.768 & 0.805755 \\
				DistalPhalanxO.C.     & 0.82          & 0.791367      & 0.822464            & 0.822464      & 0.822464      & 0.793478      & 0.7823   & 0.8152       & 0.782608696 & 0.76956522 & 0.786 & 0.826087 \\
				Earthquakes           & 0.801         & 0.81295       & 0.755396            & 0.755396      & 0.76259       & 0.755396      & 0.7482   & 0.777        & 0.741007194 & 0.74820144 & 0.755 & 0.798561 \\
				ECG200                & 0.92          & 0.91          & 0.94                & 0.95          & 0.94          & 0.93          & 0.8618   & 0.87         & 0.93        & 0.906      & 0.92  & 0.96     \\
				ECG5000               & 0.9482        & 0.948222      & 0.941556            & 0.943556      & 0.944222      & 0.940444      & 0.9454   & 0.9458       & 0.940888889 & 0.94715556 & 0.949 & 0.944222 \\
				ECGFiveDays           & 1             & 0.987224      & 1                   & 1             & 1             & 1             & 1        & 0.8165       & 1           & 1          & 1     & 0.977933 \\
				FaceFour              & 1             & 0.943182      & 0.954545            & 0.965909      & 0.977273      & 0.215909      & 1        & 0.7273       & 0.954545455 & 0.97727273 & 0.966 & 0.977273 \\
				FacesUCR              & 0.958         & 0.941463      & 0.957561            & 0.947805      & 0.926829      & 0.95122       & 0.9663   & 0.7771       & 0.971219512 & 0.96141463 & 0.98  & 0.954146 \\
				FordA                 & 0.9727        & 0.976515      & 0.948485            & 0.946212      & 0.517424      & 0.940909      & 0.941    & 0.9386       & 0.961363636 & 0.94439394 & 0.955 & 0.939394 \\
				FordB                 & 0.9173        & 0.792593      & 0.838272            & 0.830864      & 0.838272      & 0.823457      & 0.8296   & 0.8222       & 0.861728395 & 0.80506173 & 0.874 & 0.82716  \\
				GunPoint              & 1             & 1             & 1                   & 1             & 1             & 1             & 1        & 1            & 1           & 1          & 1     & 1        \\
				Ham                   & 0.781         & 0.809524      & 0.761905            & 0.780952      & 0.619048      & 0.514286      & 0.7152   & 0.7905       & 0.714285714 & 0.72571429 & 0.829 & 0.838095 \\
				HandOutlines          & 0.9487        & 0.954054      & 0.937838            & 0.948649      & 0.835135      & 0.945946      & 0.9322   & 0.9514       & 0.954054054 & 0.94243243 & 0.951 & 0.959459 \\
				Haptics               & 0.551         & 0.558442      & 0.564935            & 0.545455      & 0.600649      & 0.194805      & 0.5168   & 0.5162       & 0.548701299 & 0.52402597 & 0.542 & 0.564935 \\
				Herring               & 0.703         & 0.75          & 0.703125            & 0.734375      & 0.65625       & 0.703125      & 0.5881   & 0.6875       & 0.671875    & 0.6921875  & 0.688 & 0.71875  \\
				InsectWingbeatS.      & 0.6525        & 0.668687      & 0.522222            & 0.642424      & 0.535859      & 0.536364      & 0.6429   & 0.6177       & 0.638888889 & 0.65681818 & 0.647 & 0.835341 \\
				ItalyPowerDemand      & 0.97          & 0.963071      & 0.965015            & 0.969874      & 0.962099      & 0.971817      & 0.9703   & 0.9602       & 0.965014577 & 0.96958212 & 0.972 & 0.971817 \\
				Lightning2            & 0.8853        & 0.819672      & 0.852459            & 0.852459      & 0.754098      & 0.868852      & 0.7481   & 0.8525       & 0.770491803 & 0.75901639 & 0.82  & 0.901639 \\
				Lightning7            & 0.863         & 0.863014      & 0.821918            & 0.849315      & 0.383562      & 0.835616      & 0.7634   & 0.8493       & 0.835616438 & 0.82328767 & 0.863 & 0.849315 \\
				Mallat                & 0.98          & 0.98081       & 0.977399            & 0.975267      & 0.934328      & 0.979104      & 0.975    & 0.9326       & 0.955223881 & 0.95594883 & 0.98  & 0.975693 \\
				Meat                  & 1             & 0.883333      & 1                   & 1             & 1             & 1             & 0.8879   & 1            & 0.933333333 & 0.94833333 & 1     & 1        \\
				MedicalImages         & 0.792         & 0.798684      & 0.780263            & 0.765789      & 0.759211      & 0.789474      & 0.7958   & 0.7868       & 0.794736842 & 0.79947368 & 0.783 & 0.756579 \\
				MiddlePhalanxO.A.G.   & 0.8144        & 0.668831      & 0.655844            & 0.662338      & 0.623377      & 0.662338      & 0.5832   & 0.6364       & 0.551948052 & 0.59025974 & 0.617 & 0.668831 \\
				MiddlePhalanxO.C.     & 0.8076        & 0.841924      & 0.848797            & 0.848797      & 0.848797      & 0.835052      & 0.8535   & 0.8522       & 0.817869416 & 0.83848797 & 0.838 & 0.862543 \\
				MiddlePhalanxTW       & 0.612         & 0.603896      & 0.564935            & 0.577922      & 0.551948      & 0.623377      & 0.5502   & 0.6234       & 0.512987013 & 0.56038961 & 0.591 & 0.623377 \\
				MoteStrain            & 0.95          & 0.938498      & 0.940895            & 0.916933      & 0.9377        & 0.2           & 0.9475   & 0.8403       & 0.88658147  & 0.91461661 & 0.082 & 0.930511 \\
				OliveOil              & 0.9333        & 0.766667      & 0.966667            & 0.9           & 0.933333      & 0.9           & 0.8879   & 0.7333       & 0.833333333 & 0.91666667 & 0.9   & 0.966667 \\
				Plane                 & 1             & 1             & 1                   & 1             & 0.371492      & 1             & 1        & 1            & 1           & 1          & 1     & 1        \\
				ProximalPhalanxO.A.G. & 0.8832        & 0.887805      & 0.887805            & 0.892683      & 0.882927      & 0.892683      & 0.8497   & 0.8878       & 0.848780488 & 0.85560976 & 0.854 & 0.897561 \\
				ProximalPhalanxO.C.   & 0.918         & 0.931271      & 0.931271            & 0.931271      & 0.683849      & 0.924399      & 0.8882   & 0.9244       & 0.931271478 & 0.89896907 & 0.924 & 0.938144 \\
				ProximalPhalanxTW     & 0.815         & 0.843902      & 0.819512            & 0.814634      & 0.819512      & 0.819512      & 0.8186   & 0.8049       & 0.775609756 & 0.81658537 & 0.792 & 0.853659 \\
				ShapeletSim           & 1             & 1             & 1                   & 0.911111      & 0.888889      & 0.977778      & 1        & 0.5556       & 0.955555556 & 1          & 1     & 1        \\
				ShapesAll             & 0.9183        & 0.905         & 0.923333            & 0.876667      & 0.921667      & 0.933333      & 0.93     & 0.885        & 0.928333333 & 0.90683333 & 0.94  & 0.933333 \\
				SonyAIBORobotSur.1    & 0.985         & 0.980525      & 0.988353            & 0.978369      & 0.708819      & 0.985025      & 0.8264   & 0.8802       & 0.868552413 & 0.92246256 & 0.985 & 0.985025 \\
				SonyAIBORobotSur.2    & 0.962         & 0.972718      & 0.976915            & 0.974816      & 0.98426       & 0.976915      & 0.9248   & 0.8143       & 0.946484785 & 0.91259182 & 0.962 & 0.965373 \\
				Strawberry            & 0.976         & 0.986486      & 0.986486            & 0.986486      & 0.986486      & 0.986486      & 0.9663   & 0.9811       & 0.983783784 & 0.98135135 & 0.976 & 0.989189 \\
				SwedishLeaf           & 0.9664        & 0.9792        & 0.9772              & 0.9728        & 0.9696        & 0.9664        & 0.9655   & 0.968        & 0.9744      & 0.964      & 0.963 & 0.968    \\
				Symbols               & 0.9668        & 0.98794       & 0.9799              & 0.970854      & 0.976884      & 0.252261      & 0.9766   & 0.9719       & 0.980904523 & 0.97427136 & 0.98  & 0.982915 \\
				SyntheticControl      & 1             & 0.993333      & 1                   & 0.996667      & 1             & 1             & 0.9979   & 0.7133       & 0.996666667 & 0.99966667 & 1     & 1        \\
				ToeSegmentation1      & 0.9737        & 0.991228      & 0.969298            & 0.969298      & 0.97807       & 0.991228      & 0.9653   & 0.9167       & 0.964912281 & 0.96842105 & 0.974 & 0.982456 \\
				ToeSegmentation2      & 0.9615        & 0.930769      & 0.976923            & 0.953846      & 0.953846      & 0.976923      & 0.9553   & 0.9385       & 0.938461538 & 0.92384615 & 0.946 & 0.938462 \\
				Trace                 & 1             & 1             & 1                   & 1             & 1             & 1             & 1        & 1            & 1           & 1          & 1     & 1        \\
				TwoLeadECG            & 1             & 1             & 1                   & 1             & 1             & 1             & 0.9946   & 0.9895       & 0.995610184 & 0.99912204 & 1     & 1        \\
				TwoPatterns           & 1             & 0.99675       & 1                   & 1             & 1             & 1             & 1        & 0.5157       & 1           & 1          & 1     & 1        \\
				UWaveGestureL.All     & 0.9685        & 0.961195      & 0.856784            & 0.933277      & 0.939978      & 0.879118      & 0.9689   & 0.9375       & 0.951982133 & 0.97537688 & 0.96  & 0.893076 \\
				UWaveGestureL.X       & 0.8308        & 0.843663      & 0.780849            & 0.814629      & 0.810999      & 0.808766      & 0.8411   & 0.7074       & 0.824958124 & 0.85474595 & 0.84  & 0.805974 \\
				UWaveGestureL.Y       & 0.7585        & 0.765215      & 0.664992            & 0.71636       & 0.671413      & 0.67895       & 0.7723   & 0.7513       & 0.767169179 & 0.77398102 & 0.781 & 0.698492 \\
				UWaveGestureL.Z       & 0.7725        & 0.795924      & 0.756002            & 0.761027      & 0.760469      & 0.762144      & 0.7844   & 0.7289       & 0.764098269 & 0.79190396 & 0.788 & 0.738693 \\
				Wafer                 & 1             & 0.998378      & 0.99854             & 0.998215      & 0.99854       & 0.999027      & 0.9991   & 0.9977       & 0.998539909 & 0.99823167 & 1     & 0.99562  \\
				Wine                  & 0.889         & 0.833333      & 0.851852            & 0.87037       & 0.87037       & 0.907407      & 0.8906   & 0.6667       & 0.611111111 & 0.81296296 & 0.87  & 0.962963 \\
				WordSynonyms          & 0.779         & 0.680251      & 0.661442            & 0.65047       & 0.636364      & 0.678683      & 0.7874   & 0.685        & 0.73354232  & 0.75344828 & 0.766 & 0.592476 \\
				ACSF1                 & ---           & 0.9           & 0.96                & 0.91          & 0.93          & 0.17          & ---      & 0.78         & 0.92        & 0.886      & ---   & 0.93     \\
				BME                   & ---           & 0.993333      & 1                   & 1             & 1             & 1             & ---      & 1            & 0.993333333 & 1          & ---   & 1        \\
				Chinatown             & ---           & 0.982609      & 0.985507            & 0.985507      & 0.985507      & 0.985507      & ---      & 0.7246       & 0.985422741 & 0.98250729 & ---   & 0.988338 \\
				Crop                  & ---           & 0.74494       & 0.743869            & 0.742738      & 0.746012      & 0.740476      & ---      & 0.7559       & 0.772202381 & 0.75134524 & ---   & 0.725119 \\
				DodgerLoopDay         & ---           & 0.6375        & 0.5375              & 0.55          & 0.4625        & 0.5           & ---      & 0.4875       & 0.15        & 0.5725     & ---   & 0.6125   \\
				DodgerLoopGame        & ---           & 0.898551      & 0.876812            & 0.891304      & 0.550725      & 0.905797      & ---      & 0.6812       & 0.855072464 & 0.87318841 & ---   & 0.905797 \\
				DodgerLoopWeekend     & ---           & 0.978261      & 0.963768            & 0.978261      & 0.949275      & 0.963768      & ---      & 0.942        & 0.971014493 & 0.97463768 & ---   & 0.985507 \\
				GunPointAgeSpan       & ---           & 0.996835      & 0.996835            & 0.996835      & 1             & 0.848101      & ---      & 0.9873       & 0.987341772 & 0.99683544 & ---   & 1        \\
				GunPointMaleV.F.      & ---           & 1             & 1                   & 1             & 0.996835      & 0.996835      & ---      & 0.9937       & 0.993670886 & 0.99841772 & ---   & 1        \\
				GunPointOldV.Y.       & ---           & 0.993651      & 1                   & 1             & 1             & 0.990476      & ---      & 0.981        & 0.965079365 & 0.99111111 & ---   & 1        \\
				InsectEPGRegularT.    & ---           & 0.995984      & 1                   & 1             & 1             & 1             & ---      & 0.9719       & 1           & 1          & ---   & 1        \\
				InsectEPGSmallT.      & ---           & 0.935743      & 0.955823            & 0.927711      & 0.971888      & 0.477912      & ---      & 0.9438       & 0.9437751   & 0.97911647 & ---   & 0.919679 \\
				MelbournePedestrian   & ---           & 0.913061      & 0.912245            & 0.911837      & 0.904898      & 0.901633      & ---      & 0.3604       & 0.913899139 & 0.90438704 & ---   & 0.889299 \\
				PowerCons             & ---           & 0.994444      & 0.933333            & 0.944444      & 0.927778      & 0.927778      & ---      & 0.9389       & 0.944444444 & 0.94       & ---   & 0.977778 \\
				Rock                  & ---           & 0.92          & 0.78                & 0.92          & 0.82          & 0.76          & ---      & 0.78         & 0.8         & 0.9        & ---   & 0.78     \\
				SemgHandG.Ch2         & ---           & 0.91          & 0.866667            & 0.916667      & 0.848333      & 0.651667      & ---      & 0.7867       & 0.816666667 & 0.92683333 & ---   & 0.87     \\
				SemgHandM.Ch2         & ---           & 0.56          & 0.513333            & 0.504444      & 0.391111      & 0.468889      & ---      & 0.5244       & 0.482222222 & 0.64511111 & ---   & 0.533333 \\
				SemgHandS.Ch2         & ---           & 0.873333      & 0.746667            & 0.74          & 0.666667      & 0.788889      & ---      & 0.6644       & 0.824444444 & 0.88111111 & ---   & 0.8      \\
				SmoothSubspace        & ---           & 0.98          & 1                   & 1             & 0.993333      & 1             & ---      & 0.9933       & 0.993333333 & 0.97866667 & ---   & 1        \\
				UMD                   & ---           & 0.986111      & 1                   & 1             & 1             & 1             & ---      & 0.8264       & 0.986111111 & 0.99236111 & ---   & 1        \\ \bottomrule
		\end{longtable}
\end{landscape}
}

\begin{table*}[htb]\tiny
	\begin{center}
		\caption{Performance summary of 12 algorithms on 85 UCR2018 datasets}
		\label{tab1}
		\begin{tabular}{ccccccc}
			\hline
			& Existing SOTA        & Best:LSTM-FCN        & Vanilla:ResNet-Trans & ResNet-Trans 1  & ResNet-Trans 2  & ResNet-Trans 3  \\ \hline
			Total                & 65                   & 85                   & 85                         & 85                   & 85                   & 85                   \\
			Win                  & 5                    & 13                   & 2                          & 0                    & 2                    & 0                    \\
			Tie                  & 13                   & 10                   & 21                         & 16                   & 15                   & 19                   \\
			Lose                 & 47                   & 62                   & 62                         & 69                   & 68                   & 66                   \\
			Best                 & 18                   & 23                   & 23                         & 16                   & 17                   & 19                   \\
			MeanAcc             & 0.9001               & 0.8932               & 0.8866                     & 0.8833               & 0.8230               & 0.8077               \\
			Avg\_rank             & 6.1588               & 5.1823               & 5.3176                     & 5.9588               & 6.8294               & 6.5058               \\ \hline
			& TS-CHIEF            & ROCKET                & InceptionTime              & MACNN                 & ResNet-50 SC           & FMLA                \\ \hline
			Total                & 65                   & 85                   & 85                         & 65                   & 85                   & 85                   \\
			Win                  & 1                    & 7                    & 3                          & 9                    & 1                    & 14                   \\
			Tie                  & 8                    & 10                   & 7                          & 14                   & 6                    & 22                   \\
			Lose                 & 56                   & 68                   & 75                         & 42                   & 78                   & 49                   \\
			Best                 & 9                    & 17                   & 10                         & 23                   & 7                    & 36                   \\
			MeanAcc             & 0.8679               & 0.8814               & 0.8653                     & 0.8692               & 0.8249               & 0.8937               \\
			Avg\_rank             & 8.1470               & 6.6941               & 7.1117                     & 5.4461               & 8.2882               & 4.8176               \\ \hline
		\end{tabular}
	\end{center}
\end{table*}

\begin{figure*}[!t]
	\centering
	\includegraphics[width=6in]{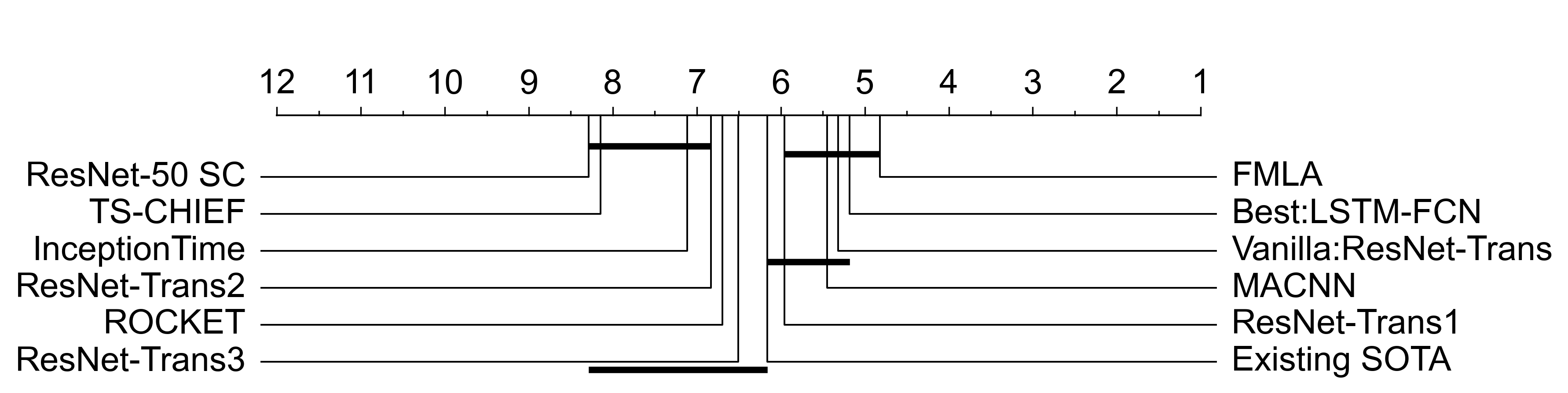}
	\caption{Results of AVG\_rank of 12 algorithms on 85 univariate datasets}
	\label{avg}
\end{figure*}

\subsection{Experimental Analysis}
To evaluate the performance of FMLA, we compare it with 11 advanced algorithms including deep learning-based algorithms and traditional ones, with experiments on 85 UCR2018 datasets conducted as shown in Tables \ref{overall} and \ref{tab1}. All these models are listed below.

- Existing SOTA: a well-known benchmark consisting of the highest accuracy on each dataset obtained by STC \cite{ruizGreatMultivariateTime2021}, HC \cite{largeProbabilisticClassifierEnsemble2019}, gRSF \cite{karlssonGeneralizedRandomShapelet2016} and mv-ARF \cite{tuncelAutoregressiveForestsMultivariate2018}.

- Best:LSTM-FCN \cite{karimLSTMFullyConvolutional2018}: a dual-network algorithm consisting of an attention-based LSTM branch and an FCN branch.

- Vanilla: ResNet-Transformer \cite{huangResidualAttentionNet2020}: a dual-network algorithm consisting of a ResNet branch and a vanilla Transformer branch.

- ResNet-Transformer 1, 2, 3 \cite{huangResidualAttentionNet2020}: three dual-network algorithms consisting of a ResNet branch and a Transformer branch with a different number of blocks.

- TS-CHIEF \cite{shifazTSCHIEFScalableAccurate2020}: a tree classifier based on the Proximity Tree algorithm.

- ROCKET \cite{dempsterROCKETExceptionallyFast2020}: an ensemble algorithm consisting of a large number of random convolutional kernels.

- InceptionTime \cite{fawazInceptionTimeFindingAlexNet2020}: an application of the Inception network in TSC

- MACNN \cite{chenMultiscaleAttentionConvolutional2021}:  an attention-based multi-scale CNN.

- ResNet-50 SC \cite{wenningerTimageRobustTime2019}: a ResNet-50 architecture with different classifiers on different datasets.

- FMLA: the proposed model in this paper.

As shown in Table \ref{tab1}, the proposed FMLA performs the best among the 12 algorithms. To be specific, it obtains the best top-1 accuracy results on 36 out of the 85 datasets and wins 22 cases. FMLA ranks first regarding the average ranking shown in Fig. \ref{avg}. Besides, it is obvious that the top five algorithms are all based on the attention mechanism, which, to some extent, proves the potential of attention. LSTM-FCN has a slightly lower performance compared with ours. The LSTM improved by attention mechanism makes LSTM-FCN acquire similar advantages to Transformer-based algorithms. However, its intrinsic disadvantage, namely being biased on the latest information, still exists. This may lose previous information in long time series, leading to weakened performance. Moreover, FMLA outperforms the 4 Transformer-based models in \cite{huangResidualAttentionNet2020}, one of which achieves the third place in the average ranking. These four models perform similarly in terms of `Best' but obtain different mean accuracy values. This means they have similar feature extraction abilities but are adaptive to datasets with different sizes based on their Params. As for the fourth algorithm in average ranking, MACNN, it applies the attention block after each multi-scale CNN block to aggregate the multi-scale information obtained, meaning that the relative importance of the features extracted by CNN blocks can be distinguished so that more useful features contribute more to the classification result. The average rank of MACNN is slightly lower than the vanilla:ResNet-Transformer model but outperforms the other 3 variants Transformer-based model.

Both Rocket and InceptionTime use multiple kernels for multi-scale feature extraction, wherein the large number of random convolutional kernels used in Rocket guarantee to extract abundant features compared with `InceptionTime'. However, both of them cannot well analyze the importance of each feature extracted without the attention mechanism. 
The `TS-CHIEF' and `ResNet-50 SC' are the worst two algorithms since they use regular methods to extract monotonous features. It is thus difficult for them to extract flexible shapelets in TSC with noise influence.

To sum up, FMLA achieves the best performance in terms of `win'/`tie'/`lose', MeanACC, and AVG\_rank. On the one hand, FMLA inferences more efficiently than the vanilla attention-based and LSTM-based algorithms according to its better parallelism and lower complexity. On the other hand, FMLA performs better than all the other attention-irrelevant algorithms as it is able to provide a better combination of local and global features through the integration of DCN and CLA blocks.

\begin{figure*}[!t]
	\centering
	\subfloat[FLOPs]{\includegraphics[scale=0.55]{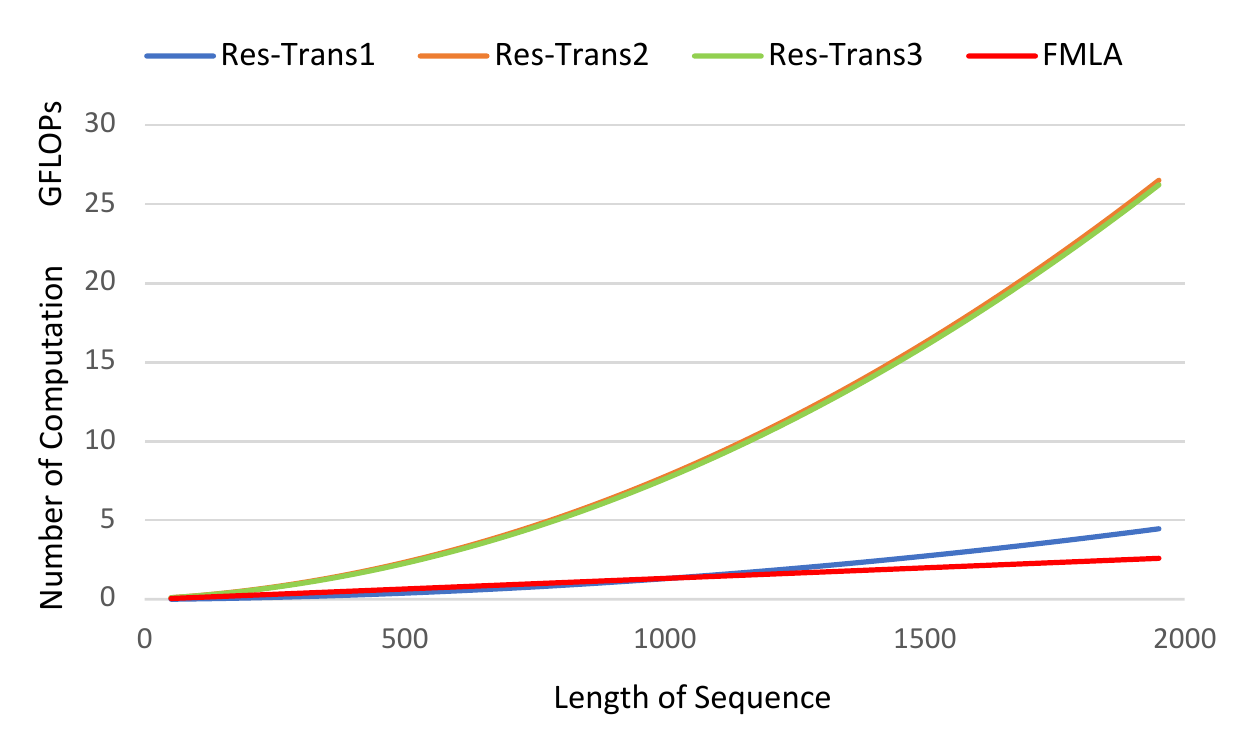}}
	\subfloat[Params]{\includegraphics[scale=0.55]{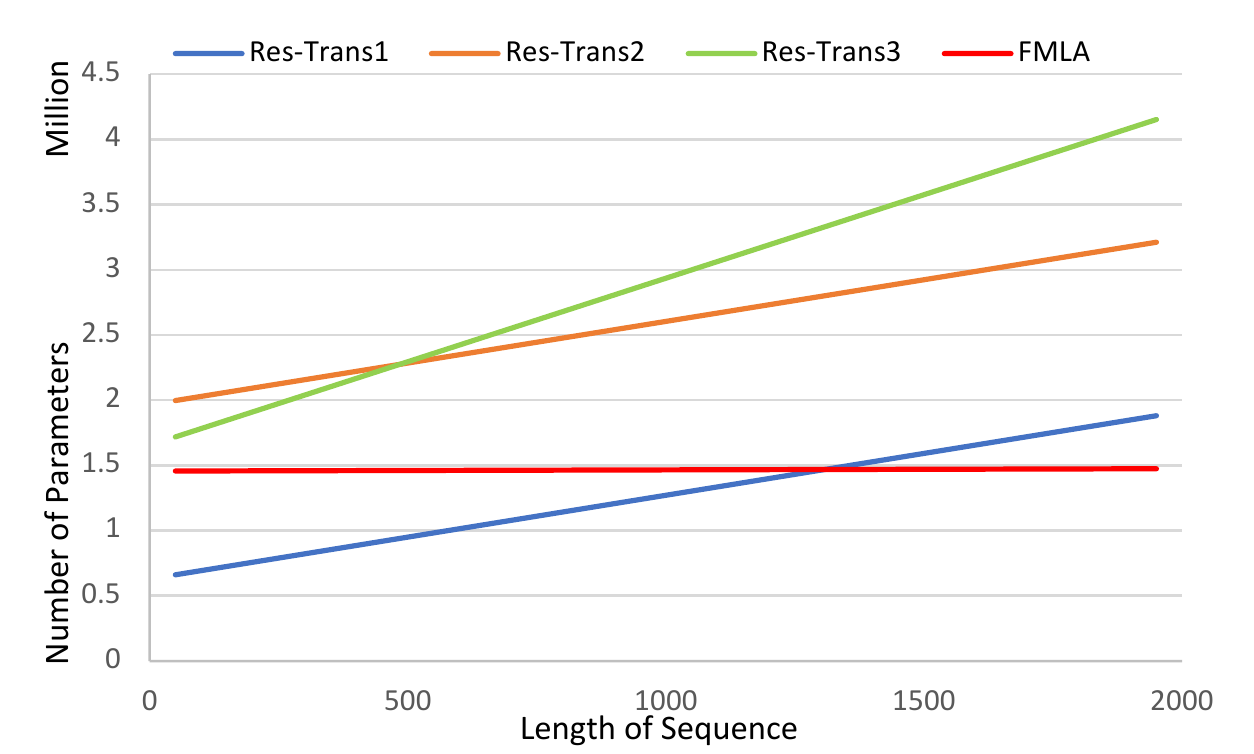}}
	
	\captionsetup{justification=centering}
	\caption{Complexity comparison in terms of FLOPs and Params}
	\label{complexity}
\end{figure*}

\subsection{Parallelism and Complexity Analysis}
LSTM-FCN \cite{karimLSTMFullyConvolutional2018} is one of the pioneering studies on TSC.  It processes a sequence token by token to obtain global representations. However, the weak parallelism causes its training time highly dependent on the length of a given series and it needs more parameters to remember more previous information. To overcome these problems, attention-based models with better parallelism have been proposed, like ResNet-Transformer \cite{huangResidualAttentionNet2020} and MACNN \cite{chenMultiscaleAttentionConvolutional2021}. Attention treats information in sequences as query-key pairs, which leads to quadratic complexity. Assume $d$ is the dimension of input vectors and $n$ is the length of the input sequence. There are mainly four processes directly related to the complexity of the vanilla attention mechanism in \cite{huangResidualAttentionNet2020} and \cite{chenMultiscaleAttentionConvolutional2021}, including the projection of queries, keys and values with $O(3nd^2)$, dot-product calculation of similarity with $O(n^2d)$, SoftMax calculation of similarity with $O(n^2)$ and weighted sum calculation of values with $O(n^2d)$.

In order to reduce the complexity of vanilla attention, we compress the length of values based on the low-level features extracted by DCN blocks and the keys are generated from the compressed values. The projections of keys in the vanilla attention are omitted so the complexity in this stage is reduced to $O(2nd^2)$. Otherwise, FMLA has a linear complexity of $O(Cnd)$, in the dot-product, SoftMax and weighted sum calculation processes. The mask mechanism is actually a broadcast operation with a complexity of $O(nd)$ and the knowledge distillation techniques are actually none-parameter operations that do not influence the complexity directly. All in all, the complexity of our model, $O(n)$, is much lower than that of the vanilla Transformer-based model, $O(n^2)$, and our parallelism is better than the vanilla attention-based models.

The FLOPs and Params of the FMLA model are compared with those of three Transformer-based models, as shown in Fig. \ref{complexity}. ResNet-Transformer2 and ResNet-Transformer3 own the same number of blocks as ours but their FLOPs increases quadratically. The ResNet-Transformer1 has fewer FLOPs in the early stage because it only uses one Transformer block. However, as the length of the given sequence grows, the FLOPs demonstrate the advantages of FMLA in efficiency. In terms of Params, all four algorithms grow linearly, where the growth mainly happens in the final classification layers. In FMLA, we squeeze the outputs of two branches and add to the results before classification without a specific classifier. Therefore, the number of parameters we used grows slower than the others with the growth of the sequence length.

\section{Conclusion}
In this paper, we propose a global-local attention model, namely FMLA, for time series classification. FMLA integrates the deformable convolution network and the collaborative attention mechanism, a mask mechanism and online knowledge distillation. The FMLA realizes global and local feature awareness, low noise interference and model redundancy reduction. Sufficient and comprehensive experiments are conducted on 85 datasets from UCR2018 and FMLA achieves the best results on 36 datasets, which outperforms all other comparing algorithms in accuracy. Based on our extensive experiments and analysis, our model can also extract features with lower complexity for time series classification compared with other deep learning algorithms.

\section{Acknowledgements}
This work was supported in part by the National Natural Science Foundation of China (No. 62172342), the Natural Science Foundation of Sichuan Province (No. 2022NSFSC0568), and the Fundamental Research Funds for the Central Universities, P. R. China.


  \bibliographystyle{plain} 
  \bibliography{refs}


%\begin{thebibliography}{00}


%\bibitem{}

%\end{thebibliography}
\end{document}